\pdfoutput=1

\documentclass[11pt]{article}

\usepackage[final]{ACL2023}

\usepackage{times}
\usepackage{latexsym}
\usepackage{amssymb}
\usepackage{subcaption}
\usepackage{array}
\usepackage{makecell}
\usepackage{amsmath}

\usepackage[T1]{fontenc}

\usepackage[utf8]{inputenc}

\usepackage{microtype}

\usepackage{inconsolata}

\usepackage{graphicx}
\usepackage{float}
\usepackage{enumitem}
\usepackage{booktabs}

\definecolor{todo}{rgb}{1,0.5,0}
\definecolor{remark}{rgb}{1,0.5,1}

\newcommand{\stricta}{STRICTA}

\newtheorem{definition}{Definition}

\newcolumntype{P}[1]{>{\centering\arraybackslash}m{#1}}

\title{STRICTA: Structured Reasoning in Critical Text Assessment\\for Peer Review and Beyond}

\author{Nils Dycke$^\dagger$, Matej Zečević$^{*}$, Ilia Kuznetsov$^\dagger$ \\
    \textbf{Beatrix Suess$^\ddagger$, Kristian Kersting$^*$, Iryna Gurevych$^\dagger$} \\ \\
  $^\dagger$ UKP Lab, ATHENE National Research Center for Applied Cybersecurity,\\ Technical University of Darmstadt \\
  $^*$ AIML Lab, Hessian Center for AI, Technical University of Darmstadt\\
  $^\ddagger$ Synthetic RNA Biology, Technical University of Darmstadt\\
  }

\begin{document}
\maketitle
\begin{abstract}
Critical text assessment is at the core of many expert activities, such as fact-checking, peer review, and essay grading. Yet, existing work treats critical text assessment as a black box problem, limiting interpretability and human-AI collaboration. To close this gap, we introduce \textbf{S}tructured \textbf{R}easoning \textbf{I}n \textbf{C}ritical \textbf{T}ext \textbf{A}ssessment (STRICTA), a novel specification framework to model text assessment as an explicit, step-wise reasoning process. STRICTA breaks down the assessment into a graph of interconnected reasoning steps drawing on causality theory \cite{pearl1995causal}. This graph is populated based on expert interaction data and used to study the assessment process and facilitate human-AI collaboration. We formally define \stricta~ and apply it in a study on biomedical paper assessment, resulting in a dataset of over 4000 reasoning steps from roughly 40 biomedical experts on more than 20 papers. We use this dataset to empirically study expert reasoning in critical text assessment, and investigate if LLMs are able to imitate and support experts within these workflows. The resulting tools and datasets pave the way for studying collaborative expert-AI reasoning in text assessment, in peer review and beyond.\footnote{Code and data are publicly available at \url{https://github.com/UKPLab/acl2025-stricta}}
\end{abstract}

\section{Introduction}
During critical text assessment, an expert evaluates a document and arrives at a verdict about the document's quality. It is a key task in many expert domains, and often serves as a cornerstone for quality assurance in the respective area \cite{lewandowsky2020debunking,johnson_stm_2018,royce2012assessment}. Critical text assessment is challenging: the expert must analyze the document from multiple perspectives, integrate background knowledge, and form coherent reasoning to arrive at a judgment. To support the experts, prior work has explored various approaches to automated text assessment in specific domains, such as document quality classification \cite{shen2019joint,maillette-de-buy-wenniger-etal-2020-structure}, peer review report generation \cite{d2024marg,yuan2022can,du-etal-2024-llms,yu-etal-2024-automated}, or automatic essay grading \cite{misgna2025survey,han-etal-2024-llm}. Yet, prior work largely treats critical text assessment as a black box, leaving a crucial question unanswered: \textit{How do the experts arrive at their judgement?} 
Answering this question is a key prerequisite for transparent and reliable human-AI collaboration in critical text assessment. It would enable comparative analysis of human decision-making, provide scaffolding for fine-grained evaluation of AI assistance, and facilitate human-AI collaboration via enhanced interpretability and human oversight \cite{luo2023reasoning, ferdaus2024towards}. Yet, the lack of formal models and datasets that capture expert reasoning during critical text assessment prevents progress in this area. Our work aims to address this gap.

\begin{figure}[t] 
  \centering
  \includegraphics[width=\linewidth]{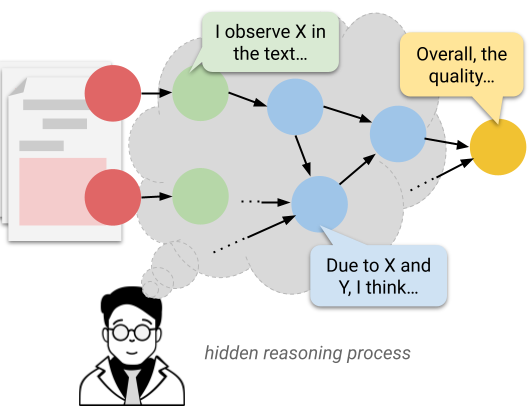}
  \caption{Structured Reasoning in Critical Text Assessment (\stricta) represents critical text assessment as a reasoning process decomposed into causally linked steps. Based on the observed text, the expert produces a verdict by filling-in each step.}
  \label{fig:overview}
\end{figure}

To formalize expert reasoning, we introduce \stricta: a specification framework that represents critical text assessment as a step-wise structured reasoning graph (\emph{workflow}), capturing how different aspects of a document contribute to the final evaluation (Figure \ref{fig:overview}). We propose a three-step process for constructing and populating workflows based on expert interactions. Since many text assessment tasks are partially subjective and allow for
multiple valid interpretations, we explore the use of structural causal models (SCMs) \cite{pearl2000models} to explicitly account for this variance during reasoning. We also investigate how workflows can be used to evaluate LLMs' ability to imitate human reasoning during critical text assessment, casting it as an abductive reasoning problem.

The assessment of scientific papers underlying the peer review process is arguably one of the most challenging document assessment tasks requiring complex reasoning. To demonstrate the use of \stricta~in practice, we therefore focus on paper assessment -- a critical text assessment task closely related to the emerging line of work on AI-assisted peer review \cite{Shah2022AnOO,Kuznetsov2024WhatCN,Goldberg2024UsefulnessOL}. Specifically, we conduct a large-scale study in biomedical paper assessment. We construct reasoning workflows based on expert interviews, and engage over $40$ biomedical researchers and students to collect workflow-based assessments for more than $20$ papers. With over $4000$ explicit reasoning steps, this unique dataset provides an empirical foundation for studying how experts reason during text assessment, and how modern LLMs can assist them in this task. Overall, we contribute:

\begin{itemize}[leftmargin=*, noitemsep]
    \item a model for \stricta~tasks rooted in causality, and a framework for its application;
    \item a unique expert-generated multi-modal structured reasoning dataset in biomedical paper assessment;
    \item insights into human influence factors during paper assessment and the potential of LLM assistants in \stricta~ applied to paper assessment.
\end{itemize}

\noindent Our analysis reveals that \textit{differences in prior knowledge} are a key source of disagreement among experts, while a paper’s \textit{writing style} has a major impact on the overall assessment. Experiments with four state-of-the-art LLMs demonstrate that LLMs are prone to error propagation, but human oversight effectively mitigates this issue. Our study connects research on causality, expert-AI collaboration, and LLMs and paves the path towards transparent and reliable AI assistance in expert domains. 

\section{Related Work} \label{sec:rw}
Our work offers a novel perspective on critical text assessment with a practical focus on research paper assessment. It directly relates to research on peer review generation and other document quality assessment domains such as fact-checking. The proposed approach is also inspired by research on abductive and diagnostic reasoning and relates to causality in NLP, which we briefly cover in Appendix \ref{asec:extended_rw}.
Existing work on text assessment trains models to predict the quality of documents such as scientific papers or Wikipedia articles using external quality labels (e.g., citation counts) \cite{shen2019joint,maillette-de-buy-wenniger-etal-2020-structure,wenniger2023multischubert}. However, these methods rely on proxy indicators without human validation and treat document assessment as a black-box problem. Our work complements these approaches by focusing on human assessments and the reasoning process to enhance interpretability.
Automatic review generation aims to produce full peer review reports for papers using specialized model architectures \cite{d2024marg,yuan2022can}, testing LLMs out-of-the-box \cite{du-etal-2024-llms,tyser2024ai,liu2023reviewergpt,yu-etal-2024-automated}, or providing isolated feedback on paragraphs \cite{chamoun-etal-2024-automated}. While these methods offer valuable insights into downstream performance in peer review, they neglect the fine-grained decision-making factors in humans and LLMs that shape the complex assessment. Our study focuses on the underlying structured reasoning process rather than generating a review report, enabling a deeper analysis.
Another related area of research is explanations in fact-checking, focused on assessing the veracity of text \cite{guo-etal-2022-survey}. Prior studies have generated structured explanations for final verdicts as rule chains \cite{10.1145/3289600.3290996}, argumentation graphs \cite{si-etal-2024-checkwhy}, multi-hop graphs \cite{jiang-etal-2020-hover}, or verification programs \cite{pan-etal-2023-fact}, focusing on short factual claims and often ignoring ambiguity  \cite{glockner-etal-2024-ambifc}. In contrast, our approach explicitly models the uncertainties induced by subjectivity and ambiguity, explores diverse decision-making factors in humans, and focuses on whole documents.

\section{Framework}

\subsection{Background: Causal Models} \label{ssec:back}
As causal models are not ubiquitous in NLP, we provide a short introduction. Causality offers a formal framework for analyzing cause-and-effect relationships, i.e., the impact of active changes to one variable (\textit{interventions}) on the likelihood of another. This work adopts Structural Causal Models (SCMs, \citet{pearl2000models}), a widely used class of causal models. For brevity, we provide a high-level overview of SCMs and their associated concepts. Appendix \ref{ss:causality_background} provides the formal definitions.

Following \citet{pearl2000models}, an SCM $\mathcal{M}$ is a tuple $(U, V, F, P_\mathcal{M})$ where $U$ are the background variables distributed by $P_\mathcal{M}$, $V$ are the modeled variables, and $F$ is the set of structural equations. Each modeled variable is given by a structural equation $v_i = f_i(\textit{pa}_i, u_i)$ where $\textit{pa}_i$ is the set of \textit{parent} variables from $V$ and $U$. The background factors $U$ add randomness to the otherwise deterministic structural equations. We represent SCMs by a causal graph. An edge from $v_1$ to $v_2$ indicates that $v_1$ is input to the structural equation of $v_2$ (the \textit{child}). 
For an intervention $do(X=x)$, we alter the SCM $\mathcal{M}$ by replacing $X$ by $x$ in $F$ resulting in $\mathcal{M}_{X=x}$. \textit{Counterfactuals} pose what-if questions on the SCM. For a query, "What value had $Y$ if $X=x$ under the circumstances $u$?" we imagine the effect of $X=x$ on $Y$ for a concrete instance counter to the observed facts. The \textit{average causal effect} (ACE) of an intervention on a binary variable $X$ is the difference of expected values for $Y$ with $\textit{ACE}(X, Y) = \mathbb{E}[Y_{x=1}] - \mathbb{E}[Y_{x=0}]$ for $Y_x(u)$ the potential response of $Y$ under $\mathcal{M}_{X=x}$. We use ACE and counterfactuals during analysis (Sec.~\ref{sec:analysis}).

\subsection{Structured Reasoning in Critical Text Assessment (\stricta)} \label{ssec:framework}
We define our Structured Reasoning in Critical Text Assessment framework to model a wide range of document assessment tasks with minimal assumptions. This framework assumes an expert extracts key information from the document as input for a series of interconnected reasoning steps. These steps culminate in a final verdict on the document's quality and provide explanations linking the text to the outcome. The reasoning steps represent the typically unobserved mental processes that influence the assessment, with specific steps and connections varying based on the problem domain. 
We formalize the \stricta~problem family using SCMs, as they directly represent causal reasoning structures and facilitate the analysis of explanatory behavior in humans and AIs. Unlike purely statistical models, SCMs allow reasoning about causal rather than merely associational relationships between reasoning steps. Thus, causal modeling is indispensable for studying decision factors during text assessment.\footnote{In Section \ref{sec:analysis}, we make use of these capabilities posing complex causal queries on the constructed SCM for paper assessment. Appendix \ref{ss:nldar_alternatives} discusses alternative formalisms.}%

\begin{definition} \label{def:nldar}
    A structured reasoning for critical text assessment problem is given by an SCM $\mathcal{M}=(U, V, F, P_\mathcal{M})$, the \emph{workflow}, with
    \begin{enumerate}[label=\roman*.,noitemsep]
        \item inputs $I \subset V$ and $I \neq \emptyset$ in natural language.
        \item final verdicts on the document with $T \subseteq V$ and $T \neq \emptyset$.
        \item reasoning components $C \subset V$ with $C \cap I = \emptyset$, $C \cap T = \emptyset$ and $C \neq \emptyset$ forming the \emph{steps} of the workflow.
    \end{enumerate}
    Furthermore, all inputs are roots of the workflow graph and there is a path that connects each step with at least one input. The final verdicts are terminal nodes in the graph. The background variables determine the level of subjectivity and noise in the task. Solving a \stricta~problem means finding the most likely values for all unobserved $E_\text{hidden}\subset C\cup T$ given $I$ and partially observed values $E_\text{observed} \subset C \cup T$ with $E_\text{hidden} \cap E_\text{observed} = \emptyset$.
\end{definition} 

\noindent Informally, a workflow describes a reasoning graph that connects an input text to a final verdict. While the assessment structure remains fixed, the answers for each step vary by instance. The reasoning task depends on the observed reasoning steps. In Section \ref{sec:llm}, we present a challenging abductive reasoning problem, providing the LLM only with the paper text and final verdict, and tasking it with reconstructing the intermediate reasoning steps. 

Notably, we assume that all problem instances in a \stricta~problem class share the same reasoning structure, with variation arising only in the graph's variables. This contrasts with prior work that constructs reasoning structures for each instance individually \cite[e.g. in fact checking][]{pan-etal-2023-fact}. A fixed reasoning structure, where each step has a well-defined, context-independent meaning, enables quantitative comparison of human reasoning factors and fine-grained evaluation of automatic methods across instances.

\subsection{Instantiating a \stricta~Problem} \label{ss:instantiate}
\stricta~represents a family of document assessment problems. Addressing a specific problem, such as biomedical paper assessment, involves the following steps: %

\noindent\textbf{Step 1. Designing the SCM's Structure:} First, we specify the steps and causal links of the workflow graph. Ultimately, this means determining the influence factors on the final verdict based on the underlying document. This can either be done automatically through \textit{causal discovery} \cite{Verma1990EquivalenceAS} or through interaction with domain experts.
The steps of the workflow can be modeled as boolean decisions, numerical scores, or text answers.\footnote{Causality-aware text representation is a hard research problem without a general solution \cite{veitch2020adapting,feder2022causal}. In our case study (Section \ref{sec:scm}), we combine boolean and natural language answers.} This process yields the graph structure and variables of the workflow.

\noindent\textbf{Step 2. Populating the SCM with Data:} 
Next, we determine the relationships between the workflow steps by estimating the SCM's structural equations. To do this, human annotators follow the workflow to assess the paper, answering each step after reading the document and forming a verdict. By enforcing the step dependencies during annotation, we ensure the assumed causal constraints.
Varying experiences and biases act as background factors. If an assessment task is subjective, background factors induce high variability in the answers. To measure the level of noise, each step and input is annotated redundantly. Using this data, we estimate the structural equations by training machine learning models on the input-output data per step. 

\noindent\textbf{Step 3. Analysis and Automation:} 
Ignoring training error, the resulting SCM from Step 2 perfectly models the human reasoning process for the given problem. By formulating causal queries on the SCM, we simulate counterfactuals and interventions to describe how inputs and intermediate steps influence the final verdict. 
Additionally, the workflow serves as a backbone for assessing new documents, either automatically or through human-AI collaboration. Since the workflow decomposes the assessment task into individual steps, various reasoning and assistance scenarios can be tested. In Section \ref{sec:llm} we demonstrate that LLMs integrate naturally with the causal graph to solve \stricta~problems reconstructing human reasoning from partial information by turning the causal graph into an LLM program \cite{Dohan2022LanguageMC}.
\section{Paper Assessment as \stricta} \label{sec:scm}

\begin{figure*}[t!]
  \centering
  \includegraphics[width=0.9\linewidth]{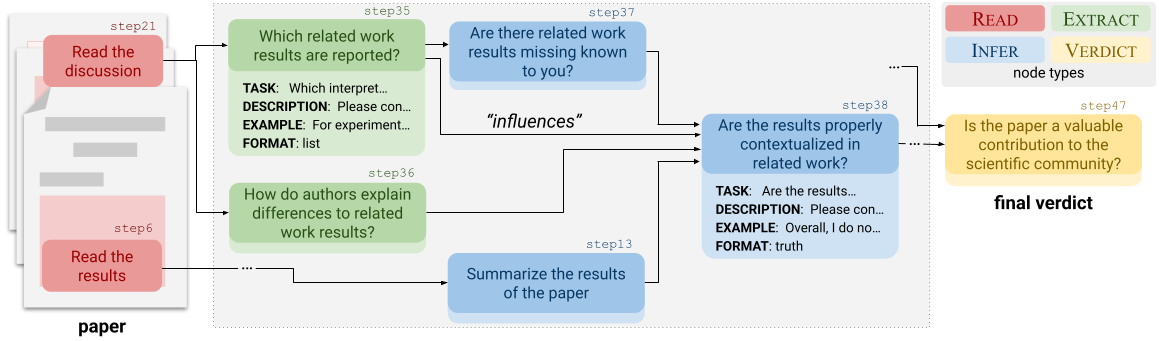}
  \caption{Paper assessment workflow (excerpt). Each node is associated with a task, description, example, and output format. "\ldots"  means there exists a path.}
  \label{fig:workflow_excerpt}
\end{figure*}

Paper assessment is a critical step in scholarly peer review and can substantially benefit from AI assistance \cite{Kuznetsov2024WhatCN}. We apply the proposed \stricta~framework to biomedical paper assessment. In this task, a reviewer is given the paper and arrives at an overall quality verdict through a series of unobserved reasoning steps. The output of this task is not a full peer review report but the "thought process" underlying it. This framing models the paper assessment process as supported by our expert interviews (see below) and establishes a setting for human-AI collaboration with full human agency over the process. We focus on biomedical paper assessment because of two main reasons: first, our expert interviews below show that paper assessment in biomedicine follows relatively consistent patterns across sub-communities, increasing the flexibility and scope of the resulting assessment workflow. Second, most studies in NLP for peer review focus exclusively on machine learning and NLP domains \cite{staudinger-etal-2024-analysis}; we contribute to bridging the gap towards other less-represented scientific domains \cite{Kuznetsov2024WhatCN}.
Following Definition \ref{def:nldar}, a workflow for paper assessment describes the mental model of what constitutes a high-quality paper, connecting observable text to the final verdict. We construct the workflow, i.e,. the SCM, reflecting this mental model, populate it through a human annotation study, and analyze human reasoning patterns following the \stricta~framework.

\subsection{Designing the SCM} \label{ss:design_workflow}
We derive the workflow for biomedical text assessment based on expert interviews, followed by an independent validation step, opting against the automatic causal discovery of the workflow from data such as peer review reports. Hereby, we ensure close alignment to the true human reasoning process and ensure the quality of the workflow since expert-elicited SCMs are commonly held as the gold standard \cite{zanga2022survey,zhang2023explainable}. Following prior work on modeling complex systems with expert-elicited SCMs \cite{rodrigues2022reflection,ashdown2024development}, we derive a \textit{best-effort} SCM that is agreeable with the mental paper quality model of domain experts while accounting for subjectivity via the SCM's background variables. 
We separately interview two experienced biomedical researchers (postdoc and professor) to understand the criteria for assessing a scientific article in this field. We aggregate and structure their responses to form a step-wise reasoning graph (detailed below), which we revise together with the two experts. Finally, we validate the graph with three PhD students from the same field in a formative study where we ask them to assess a biomedical article using the reasoning graph. Appendix \ref{ssec:interviews} provides details.

The resulting workflow has $45$ interdependent paper quality assessment steps terminating in a final verdict on the paper quality. Fig.\ref{fig:workflow_excerpt} shows an excerpt of the workflow. Based on the expert interviews, there are three key activities during assessment:
\begin{itemize}[leftmargin=*, noitemsep]
    \item for \textsc{read} steps, humans read a portion of the paper in detail.
    \item for \textsc{extract} steps, humans distill information from the text, e.g., listing the results in a figure.
    \item for \textsc{infer} steps, humans reason over prior answers to combine or reduce information into assessments, e.g., "Are the results plausible?".
\end{itemize}
We assign each workflow step one of these activities, a task prompt, a name, detailed instructions, and an example (see Appendix \ref{ssec:biomed_scm}).
The variable ranges are up to design choice. Paper assessment involves both text and figures, necessitating a multimodal representation of inputs. In our formative study, we observed that while some steps have boolean values (e.g., "Are the experimental results plausible given the related work?"), humans often provide detailed, unstructured answers. Consequently, we use textual representations for variables, supplementing them with boolean responses for \textsc{infer} nodes and image inputs for \textsc{read} steps.

\subsection{Populating the SCM} \label{sec:data_collection}
To estimate the structural equations of the SCM from data, we \textbf{collect a dataset of human reasoning steps} following the causal graph structure. We guide the participants through the workflow to assess a range of papers while redundantly annotating the same paper for the estimation of noise. 

\noindent\textbf{Source Data} We use openly licensed biomedical manuscripts from bioRxiv\footnote{\url{https://www.biorxiv.org/}} filtered by recency (06/2021 to 06/2023) and expertise keywords and stratified by topic, world region, and publication status (number of revisions and acceptance status at a peer-reviewed venue). The resulting sample of $200$ papers is filtered manually by an expert annotator to ensure high expertise and is subject to subselection by annotator preference, resulting in $22$ papers. See Appendix \ref{ssec:underlying_papers} for more details.

\noindent\textbf{Tools and Guidelines} The annotation study was conducted in two stages, \textsc{junior} and \textsc{senior}, based on participants' academic expertise. Both stages used the same guidelines, setups, and annotation interfaces. We converted the graphical workflow into a sequential questionnaire, ordered topologically and matching the paper structure. This ensured that prior answers required as inputs were available at each step. The CARE annotation platform \cite{zyska-etal-2023-care} was customized to guide annotators through the assessment. Each step included the reasoning prompt, an example, and relevant prior answers or paper text as input. Participants received guidelines and a training session before the study (see Appendix \ref{ssec:annotation_study}).

\noindent\textbf{Annotation Study}  In the \textsc{junior} stage, the annotation setup was validated within a university course, where $36$ postgraduate students in the biomedical field participated voluntarily. After minor changes to the guidelines and annotator training, the \textsc{senior} stage involved $8$ further doctoral and postdoctoral researchers in the biomedical field, as well as $2$ postgraduate students who received dedicated training in biomedical paper assessment.\footnote{We report the annotator compensation in Section \nameref{sec:ethics}.} After the study, we post-processed the \textsc{junior} data to filter out low-quality items (see Appendix \ref{ssec:annotation_study}). 

\noindent\textbf{Dataset} The resulting dataset consists of a \textsc{junior} subset with $4$ annotated papers and a \textsc{senior} subset covering $18$ papers. At least three annotators have annotated each paper and workflow step; $11$ papers have annotations by five different annotators. The amount of annotations per paper aligns with prior work on measuring human variability in a task \cite{giulianelli2023variability}, which we rely on during analysis in Section \ref{sec:analysis}. Additionally, we validate that the dataset size is sufficient for the purposes of our study using statistical analysis and report the details in Appendix \ref{ssec:underlying_papers}.
The resulting dataset uniquely supports the empirical study of human decision-making and reasoning grounded in text. Our analysis provides a template for examining any \stricta~problem, demonstrating the framework's versatility. Basic statistics for the dataset are presented in Table \ref{tab:quantities}.

\begin{table}[t] %
\centering
\small
\begin{tabular}{r c}
\toprule
\multicolumn{2}{c}{\textbf{Dataset Statistics}}\\
\midrule
\textbf{\# step answers} & $4371$ \\
\textbf{\# workflow executions} & $93$ \\
\textbf{\# papers} & $22$ \\
\textbf{vocabulary} & $7962$ \\
\textbf{\# tokens} & $240110$ \\
\textbf{avg. \# tokens per answer} & $60.04$\\
\midrule
\textbf{Krippendorff's $\alpha$} & $0.424$ \\

\bottomrule
\end{tabular}
\caption{Dataset statistics (\textsc{senior} + \textsc{junior} subsets). Krippendorff's $\alpha$ calculated on boolean nodes.
}
\label{tab:quantities}
\end{table}

\noindent\textbf{Boolean Decision SCM}
To support the analysis, we construct a condensed version of the workflow that consists only of the boolean variables while maintaining causal ancestry relations. Working on Boolean variables makes standard causal estimation tractable. We approximate structural equations with Gaussian Process classifiers \cite{rasmussen2006gps}. Appendix \ref{ssec:factors} provides details.

\subsection{Analysis} \label{sec:analysis}

To assist humans in \stricta, we first study how they perform the task. Using the populated SCM, we analyze expert decision-making in biomedical paper assessment empirically. While the workflow steps are fixed, answers vary across annotators. We evaluate the human answer variation in the workflow, indicating steps benefiting from AI assistance. 

\begin{figure}[t]
  \centering
  \includegraphics[width=\linewidth]{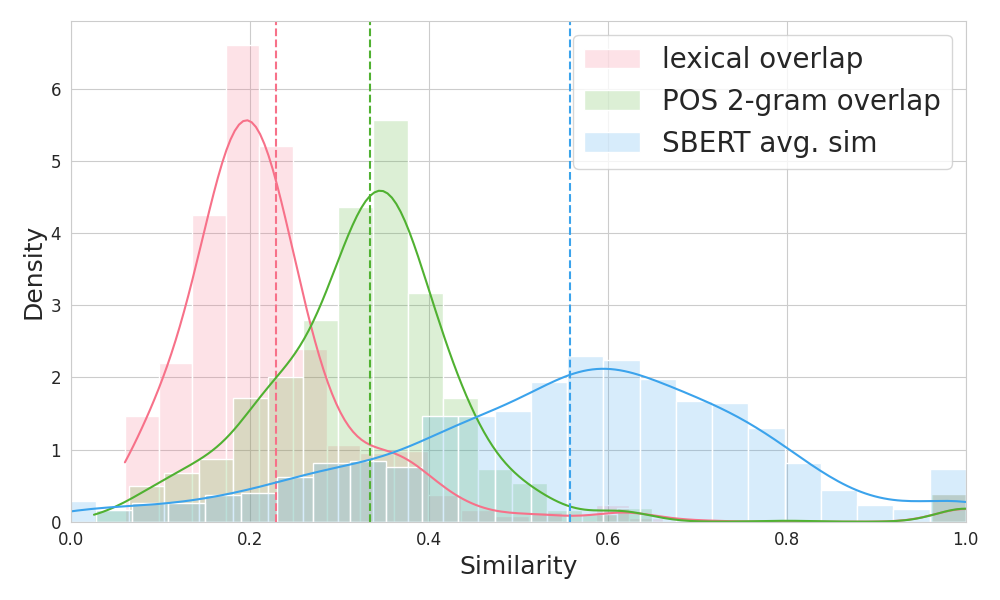}
  \includegraphics[width=\linewidth]{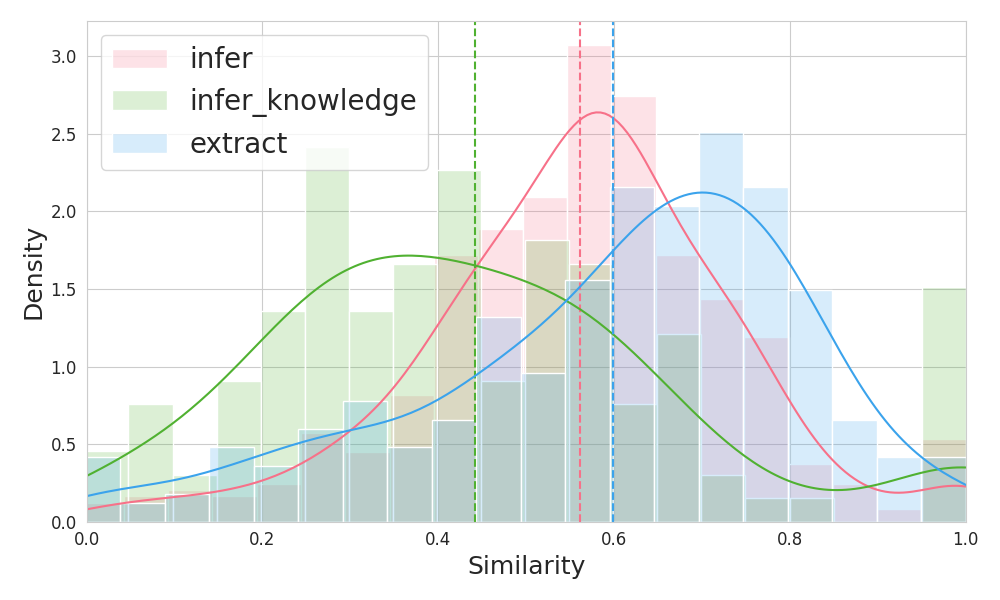}
  \caption{Average lexical, syntactic, and semantic similarity per step and paper (top) and per node type (bottom). Vertical lines are mean values.}
  \label{fig:variation}
\end{figure}

\noindent\textbf{Answer Variability} To investigate the variability in the boolean decisions, we compute the inter-annotator agreement on the steps in the boolean decision SCM. Krippendorff's $\alpha$ per step and across papers is $0.42$, indicating moderate agreement. 
This is in line with agreement levels reported for peer review ratings \cite{bornmann2011scientific,dycke2021assisting} and typical for subjective annotation tasks.
To quantify variability in the natural language answers, we follow \citet{giulianelli2023variability} and measure the noise induced by background factors in terms of the lexical variability (using lemma overlap), the syntactic variability (using part-of-speech bi-gram overlap), and semantic variability (using cosine similarity on SBERT embeddings \cite{reimers2019sbert}). We compute the metrics between all annotators per step and paper, and then take the average.
Figure \ref{fig:variation} (top) shows the answer variability distribution for all steps. Notably, the low lexical and syntactical similarity suggests that annotators employ their individual style and register and do not use template language, as has been previously observed for human-generated explanations \cite{camburu-etal-2020-make}. The semantic similarity, on average, lies at $0.55$, similar to open-ended text production tasks, such as story generation \cite{giulianelli2023variability}. Overall, there is a significant overlap between human responses, but they do raise different points. The impact of background variables is reasonably low, suggesting that no important factors were left unmodelled in the SCM.

\noindent\textbf{Sources of Variation}
We hypothesize that answer variability is not uniform across the workflow. Figure \ref{fig:variation} (bottom) shows the semantic similarity distribution for the different node types (\textsc{infer} and \textsc{extract}) and inference steps that explicitly invoke the annotators' background knowledge (\textsc{infer-knowledge}), such as \textit{"From your experience, which plot properties are common for the given plots?"}.
Answers to the steps based on background knowledge deviate the most, whereas \textsc{extract} steps are most similar. The summarization-like extraction steps invoke high consistency. \textbf{As \textsc{infer-knowledge} steps heavily depend on the individual's knowledge, prior experience is one major background factor leading to variability.} Considering variability as a proxy for step difficulty, AI support during paper assessment on background knowledge shows great potential.
Appendix \ref{ssec:variability_details} provides additional analysis, including the variability propagation in the graph.

\noindent\textbf{Impact of Decisions}
The previous analysis focuses on the background variables. We now turn to the modeled variables of the SCM. We simulate $200$ samples using the Boolean decision SCM trained on human data. Using average causal effect estimation (ACE) (Section \ref{ssec:back}), we compare the effect of interventions on different steps on the final verdict. For each node $x$ with a path to the final node $y$, we compute $\textit{ACE}(x, y)$.
Table \ref{tab:ace} lists the ACEs for selected steps. \texttt{step33} (Alignment of Conclusions and Research Questions) has the highest positive impact on the final verdict. In other words, if the findings align well with the posed questions, the final verdict turns more positive.
However, \textbf{the clarity and style of writing also play an important role} (\texttt{step48}). As the ACE directly measures the strength of causation (not an associational relation), we can conclude that annotators show a positive bias towards well-phrased papers regardless of the scientific content.
This is consistent with findings in scholarly peer review \cite{lee2013biasPR}. This result also shows that the style and writing of the paper directly influence decision making and not only lead to higher variability due to different interpretations of potentially unclear or ambiguous paper contents. This confirms our design decision to include the style of writing as an explicit confounding factor in the workflow.

\noindent\textbf{Counterfactual Analysis}
Finally, we explore the use of the SCM for counterfactual analysis. How would one need to change the underlying paper to turn the final verdict positive? Focusing on cases where annotators gave a negative verdict ($25$ instances on $12$ papers), we examine counterfactual interventions. As studying all possible $2^{18}$ interventions is intractable ($18$ steps in the boolean decision SCM), we focus on steps related to figure quality ($5$), which we call $G$. For all possible interventions on $G$, $C = [f_x=i_x: x \in G]$, we determine $P_{M_C}(t=\textit{True})$ for the final verdict $t$.
In $40\%$ of the cases, the annotator would have maintained their negative verdict. In the remaining $60\%$, an intervention on a single step -- the alignment between the figures and the text discussion (\texttt{step12}) -- would have changed the overall verdict. In other words, the revision of figures would have frequently improved the paper. Overall, this illustrates the impact of specific paper traits on the final verdict.

\begin{table} 
    \centering
    \small
    \renewcommand{\arraystretch}{1.3}
    \begin{tabular}{P{6cm}l}
         \toprule
         \textbf{Step Prompt} & \textbf{ACE} \\
         \midrule
         Considering the correct conclusions and the paper's research questions, do the conclusions fully answer the research questions?  (\texttt{step33}) & $0.37$ \\
         Overall, do the conclusions appear relevant (i.e. in-scope of the paper and its scientific field)? (\texttt{step46}) & $0.20$ \\
         Is this paper written clearly and concisely? (\texttt{step48})  & $0.20$ \\
         \midrule
         Is this a method paper? (\texttt{step4}) & $0.02$ \\
         Are the provided figures the common choice for the given methods? (\texttt{step19})& $-0.01$ \\
         Is the result reporting overall adequate for the given methods? (\texttt{step20}) & $0.00$ \\
         \bottomrule
    \end{tabular}
    \caption{Average causal effects (ACE) of workflow steps on the final verdict on the paper. We list nodes with the highest (top) and lowest (bottom) absolute effect.}
    \label{tab:ace}
\end{table}

\section{LLM Assistance} \label{sec:llm}
The \stricta~framework decomposes complex document assessment tasks, opening new opportunities for targeted assistance and interpretable AI. We investigate how LLMs can support \stricta, by testing their abilities on the collected dataset. Following Definition \ref{def:nldar}, the exact reasoning task depends on the data provided to the LLM. For our experiments, we focus on an abductive setting where the final verdict is given and all intermediate reasoning steps need to be reconstructed based on the paper.\footnote{We experiment with multiple reasoning settings in Appendix \ref{ss:extended_results}. The results are consistent.} This models the case where humans have full agency over the verdict but receive assistance to find a valid explanation for it. Specifically, we evaluate LLM alignment with human answers on intermediate steps and in a simulated assistance scenario. We highlight that our goal here is not to introduce a new NLP method, but to explore the boundaries and opportunities of existing LLMs when applied to \stricta~problems. Nonetheless, our experiments shine a new light on LLMs for causal reasoning, and suggest that a fixed reasoning structure may help address some of their inherent limitations \cite{causalparrots}.

\noindent\textbf{Setup}
We split the dataset into $20$ test papers ($4089$ answers) and reserve $2$ papers ($282$ answers) for prompt development. We experiment with LLama 3 (8B, 32k context window) \cite{llama3modelcard}, Mixtral \cite{jiang2024mixtral}, GPT-3.5 turbo \cite{brown2020language}, and GPT-4o \cite{OpenAI_GPT4_2023}. 
Appendix \ref{ssec:llm_configs} lists details on the LLMs and prompts. All prompts are designed zero-shot. 
We assess LLM outputs using automatic text similarity metrics to measure alignment with human answers across multiple dimensions. While being imperfect \cite{sottana-etal-2023-evaluation}, these metrics capture individual alignment in a human-AI collaboration setting, consistent with prior work on peer review generation comparing automated and human-written reviews \cite{du-etal-2024-llms}. We use BERT-F1 \cite{bert-score} for semantic similarity, SummaC \cite{summac_score} and TRUE \cite{honovich-etal-2022-true-evaluating} for factual alignment, and F1 score for boolean decisions. Each annotator's workflow execution is treated as an individual problem instance, with similarity metrics computed per annotator and averaged across all instances.
For validation, we complement automatic metrics with a human evaluation on a subset of the data (Appendix \ref{ss:human_eval}). Results show moderate agreement among humans -- typical for challenging tasks -- and a moderate but significant correlation between metrics and human judgment (between $\rho=0.37$ and $\rho=0.54$ for BERT-F1, and $\rho=-0.22$ and $\rho=0.55$ for SummaC, depending on the step type) at similar levels as related automatic evaluation metrics \cite{fu-etal-2024-gptscore}. Yet, these findings highlight the need for more metric development in the future for a targeted evaluation of individual steps.

\noindent\textbf{LLM Structured Reasoning}
The target task involves answering each step in the workflow to ensure that the given verdict is validly derived from the given paper text. We define a baseline LLM architecture for this process. First, we use the workflow from Section \ref{sec:scm} as a scaffold, running each LLM as a program \cite{Schlag2023LargeLM,Dohan2022LanguageMC}. The LLM is called step-by-step, with the answers from parent steps provided as inputs, thereby enforcing the workflow's reasoning structure. However, at this stage, the responses are not conditioned on the verdict being explained. To address this, in the second step, we adopt the self-refinement paradigm \cite{madaan2024self}, incorporating previous model responses, the graph structure encoded with incident encoding \cite{fatemi2024talk}, and the true verdict to refine the answers. We design tailored prompts for each workflow step type (\textsc{extract}, \textsc{infer}) and for the feedback and refinement stages. We terminate the feedback-refinement loop after one iteration.
For \textsc{read} steps, the input consists of the paper text, with figures and tables replaced by annotator descriptions for language-only LLMs. For GPT-4o, we provide figures and tables as images.
Table \ref{tab:performance} (top) summarizes the performance of the LLM program. We estimate the human baseline using leave-one-out evaluation, where each annotator's answers for a paper are compared to those of their peers, averaged across all comparisons. This approach provides a conservative lower-bound estimate of human performance.\footnote{An accurate estimate would require humans to re-annotate steps using inputs from their fellow annotators.} We calculate the F1 score based on individual annotator responses.
The GPT models and Mixtral achieve high scores on factual consistency (SummaC and TRUE), marginally above the human baseline. However, BERT-Score ranks LLMs consistently lower. The decision-making performance on the Boolean nodes lies below the average human performance by a large margin. We manually examine the discrepancy between high factual and low decision alignment in Appendix \ref{ssec:llm_extendedn_experiments}, showing that humans reserve negative decisions for extreme cases and focus their answers on a minimal set of facts relevant to the individual assessment. However, LLMs tend to restate inputs, split up facts into small components, and make harsher decisions. These factors might affect the factual alignment metrics, favoring the seemingly more extensive and specific LLM answers.  This is also supported by the relatively low factual alignment of human answers to \textsc{extract} steps with the original paper text compared to the more extensive LLM responses (details in Appendix \ref{ss:extended_results}).
While LLMs generate responses with high factual overlap compared to the average annotator, they underperform in decision-making. This indicates that LLMs weigh factors differently than humans during paper assessment, highlighting the risks of using LLMs without human oversight in \stricta~applications.

\noindent\textbf{Simulating Human-LLM Collaboration}
The \stricta~framework allows exploring simulated human-AI collaboration scenarios. Here, we assume the human oversees and corrects the LLM at each step, providing the input as human-generated answers from parent steps or the paper for \textsc{extract} steps. We exclude steps based on figures and tables to compare all LLMs. This setup evaluates the ability of LLMs to adapt to the individual's stepwise reasoning. We exclude self-refinement to focus on model performance at individual steps. Table \ref{tab:performance} (bottom, ${io}$-condition) shows the results. All LLMs demonstrate a notable performance increase, effectively adapting to the annotator’s perspective based on their step-by-step inputs and producing more aligned responses. We further investigate the reasons for these performance differences, focusing on error propagation, in Appendix \ref{ss:extended_results}. Overall, our results show the potential of human-AI collaboration for \stricta, but further research is needed to integrate this approach with the SCM of a \stricta~problem. Promising directions to mitigate error propagation include Graph-of-Thought \cite{besta2024graph} and similar architectures \cite{yao2024tree,besta2024demystifying}, which enable backtracking during reasoning.

\begin{table}[t]
    \centering
    \resizebox{\linewidth}{!}{%
    \begin{tabular}{ccccc}
    \toprule
        & \textbf{BERT-F1} $\uparrow$ & \textbf{SummaC} $\uparrow$  & \textbf{TRUE} $\uparrow$ & \textbf{F1} $\uparrow$\\
        \midrule
        human$^{*\dagger}$ & $\textbf{0.799}_{\pm.06}$ & $-0.151_{\pm.30}$ & $0.151_{\pm.27}$ & $\textbf{0.801}$ \vspace{2mm} \\      
        
        Llama3 Prg$^{\dagger}$ & $0.752_{\pm.10}$ & $-0.274_{\pm.36}$ &  $0.098_{\pm.30}$ &  $0.170$\\
        Mixtral Prg$^{\dagger}$ & $0.761_{\pm.09}$ & $\textbf{-0.149}_{\pm.26}$ & $0.120_{\pm.32}$ & $0.559$ \\
        GPT3.5t Prg$^{\dagger}$ & $0.759_{\pm.10}$ & $-0.178_{\pm.35}$ & $\textbf{0.163}_{\pm.37}$ & $0.531$ \\
        GPT4o Prg$^\dagger$ & $0.780_{\pm.07}$ & $-0.186_{\pm.30}$ &  $0.139_{\pm.35}$ & $0.720$ \\
        
        \midrule  
        \midrule
        majority$^{io}$ &  & & & $0.854$\\
        human$^{*io}$ & $0.799_{\pm.06}$ & $-0.158_{\pm.29}$ & $0.150_{\pm.27}$ & $0.801$ \vspace{2mm} \\
        Llama3$^{io}$ & $0.786_{\pm.07}$ & $-0.141_{\pm.30}$ & $0.145_{\pm.35}$ & $0.657$ \\
        Mixtral$^{io}$ & $0.794_{\pm.07}$ & $\textbf{-0.077}_{\pm.27}$ & $0.161_{\pm.37}$ & $0.822$ \\
        GPT3.5t$^{io}$ & $\textbf{0.805}_{\pm.07}$ & $-0.125_{\pm.34}$ & $\textbf{0.214}_{\pm.41}$ & $0.789$ \\
        GPT4o$^{io}$ & $0.795_{\pm.07}$ & $-0.094_{\pm.28}$ & $0.194_{\pm.40}$ & $\textbf{0.876}$ \\
        \midrule
        \midrule
        GPT4o$^\S$ & $0.776_{\pm.07}$ & $-0.154_{\pm0.28}$ & $0.188_{\pm.39}$ & $0.828$  \\
        \bottomrule
    \end{tabular}
    }
    \caption{Performance under varying conditions. $\cdot^\dagger =$ LLMs as a program. $\cdot^{io} = $ with human oversight. GPT4o$^\S$ does not use the graph. We average over steps and papers, with the standard deviation in subscript. SummaC lies in $[-1,1]$, the other metrics in $[0,1]$.}
    \label{tab:performance}
\end{table}

\section{Transfer and Future Work} \label{sec:transfer}
We design \stricta~as a general domain-independent framework for critical text assessment.
To further validate the application to other tasks besides the case study, we additionally test the generality of the \stricta~framework in a transfer experiment to paper assessment in the natural language processing domain. We adapt the biomedical assessment workflow for the NLP domain with input from two experts (PhD and Postdoc in NLP) and conduct a formative study among PhD students. We provide the detailed results in Appendix \ref{ssec:transfer}.  Overall, participants found the workflow appropriate, supporting the notion that the framework can be transferred to other domains with reasonable effort.  Since the scope of this transfer study lies in the validation of the adapted workflow, we do not undertake quality assurance measures, making the output dataset not suitable for evaluation or analysis. 
Additional validation of \stricta~ for other text assessment tasks such as essay grading, fact checking, or Wikipedia article quality assessment is important future work. This entails performing the three steps of the \stricta~framework (Section \ref{ss:instantiate}): (1) defining a workflow, (2) collecting data, and (3) using the resulting SCM and data for analysis and evaluation. Following up on our assistance experiments, humans and LLMs can be employed to jointly generate data \cite{li-etal-2023-coannotating} for speed-up. To illustrate the process of transferring \stricta~to other text assessment tasks, we describe how to instantiate \stricta~for automatic essay scoring in Appendix \ref{ss:stricta_for_aes}.

\section{Conclusion}

This paper introduced \stricta -- a novel specification framework for a family of structured reasoning problems over documents. We proposed a causality-based model, a practical guide, and a toolkit to help define and analyze such problems. Our novel, unique expert reasoning dataset instantiated the framework for biomedical paper assessment. Our experiments demonstrated the versatility of our framework for analysis and LLM assistance. Bridging the research in causality, human-AI collaboration, and LLMs, our study paves the path for more explainable, controlled, and grounded AI assistance for assessment tasks.

\section*{Limitations} \label{limitations}
\paragraph{Generalization and Domain Transfer}
Our framework for \stricta~problems is designed to be general and applicable to any text assessment task. We demonstrate its application and share the insights from our case study on biomedical paper assessment. As peer review is one of the most challenging text assessment domains requiring deep expert reasoning and observing ambiguity and subjectivity, we deem it as most appropriate for the purpose of our study. We elaborate on the domain transfer and generality of \stricta~in Section \ref{sec:transfer}.

\paragraph{Framework}
At the core of our framework lies the definition of \stricta~problems via a structural causal model. Here, we assume that the final verdicts and reasoning steps arise when consuming the input text and do not exist on their own. This perspective is subject to philosophical discussion \cite{roland1977death}. In practice, this means we assume that a text is consumed, causing the verdict (e.g., "the paper is a valuable contribution.") and that the reasoning steps in between give rise to this verdict. However, one might argue that there should be an objective verdict; for example, for paper assessment, a paper is either objectively correct or not. While we acknowledge these points, we emphasize that the practical application of our framework in the case study supports our design choices in that text assessment problems give rise to variance in the final verdict and in that the expert interviews support the directionality of causation that underlies our framework. 

\paragraph{Case Study}
The case study of this paper focuses on the biomedical domain. The above transfer experiments in Section \ref{sec:transfer} show that the developed SCM is generally relevant to paper assessment in other domains. Yet, the findings on the factors that shape human text assessment are limited to the biomedical domain; researchers from other fields might put a different focus during the assessment; for example, the contribution of datasets and open resources is not considered in the biomedical paper assessment workflow, but is an important criterion in the assessment of NLP papers. 

The data collection aims to cover a wide range of biomedical papers from various institutions and world regions to avoid biases toward a particular style of writing, scientific agenda, or scientific sub-community. However, some topical bias is inevitable to ensure relevance to the annotators' expertise.
We ensured high redundancy during annotation to avoid biasing the dataset to the idiosyncrasies of individual annotators. Yet, the given sample of papers and the annotator pool might still introduce a certain level of bias to the dataset. We highlight that the collected dataset is intended for research on reasoning during peer review across human subjects; it is not suitable for comparing behavioral patterns between groups, e.g. by varying levels of experience, which would require a different experimental design where annotators of \textit{different} groups annotate the exact same papers.

\paragraph{LLM Assistance Experiments}
In the evaluation of LLMs as assistants for structured reasoning, we rely on a range of automatic metrics for text generation. As our findings suggest, these metrics might not cover all aspects of a suitable answer in the workflow; for instance, they seem to favor more extensive answers, putting efficient but concise responses at a disadvantage. Likewise, the human baseline is a conservative estimate of true human performance; it is important to interpret our findings accordingly. Automatic metrics that consider the reasoning context of an answer, similar to the evaluation of chain-of-thought reasoning in LLMs \cite{hao2024llmreasonersnewevaluation}, are a promising direction of research to address these issues. We also conduct a human evaluation on a subset of the data to validate the automatic metrics. 
Structured reasoning offers built-in interpretability and facilitates human-AI collaboration by breaking down problems into smaller steps \cite{luo2023reasoning, ferdaus2024towards}. The findings from our simulation study on human-AI collaboration support this principle. However, an important direction for future research is to investigate how LLMs and humans \textit{interact} in real-world paper assessment scenarios. We perform a small-scale user study to validate that \stricta-based assistance is practically useful during paper assessment (see \ref{ssec:paper_diagnosis_discussion}). However, to validate these findings, more and large-scale, dedicated user studies and performance metrics focused on human-computer interaction are needed, which lie beyond the scope of this work.

\section*{Ethics} \label{sec:ethics}
This study has been approved by the university's ethical board under the proposal number \texttt{EK 41/2024}.
All senior annotators and experts participating in the study received adequate compensation within their regular employment in the research project. For the junior annotators, the workflow answers were collected as part of a university course aiming to teach students critical reading skills. The participation in the annotation study did not impact the students' grading in the underlying course. All annotators could choose freely to reject participation in the study, and all human data were collected based on explicit consent. The annotators involved in the human evaluation received $10.40$\pounds~per hour.

We explore the use of LLMs to automatically assess biomedical papers. This technology has potential dual uses in peer review. A malicious party could use the LLM to skip the manual paper assessment process, which arguably requires the most effort during peer review. This is clearly problematic because LLMs cannot entirely substitute a human researcher, leading to incorrect or biased peer reviews. We stress that the output of such a system would be the answers to the reasoning steps and not a full peer-review report, which also covers questions to the authors or suggestions for improvement. We argue that the risk of misuse of this technology is relatively low. At the same time, the benefit of enabling new research on the reasoning underlying peer review and using paper assessment workflows as a teaching or assistance device for reviewers is considerable, and we believe that the benefits of the proposed technology outweigh the risks.

Underlying our dataset is a set of publicly available papers with open licenses (CC-BY, CC0, ...) that allow the redistribution and extension of the contents. However, as we publish subjective assessments of the papers by our study participants, this could potentially harm the reputation of the publication authors when they are assessed unfairly. We highlight that this is not the intended use of our dataset, which is exclusively meant for research on \stricta. We partially mitigate this risk by having at least three different researchers make annotations per paper: as a result, the collected dataset shows an opinion spectrum rather than a singular verdict on the paper. Hereby, we deem the risk of harming authors to lie below the day-to-day risk for reputational damage in regular scientific publishing; peer reviews are commonly released to the public, and papers are discussed (and criticized) publicly, for instance, on social media and in subsequent publications.

\section*{Acknowledgments}
This research work has been funded by the German Federal Ministry of Education and Research and the Hessian Ministry of Higher Education, Research, Science, and the Arts within their joint support of the National Research Center for Applied Cybersecurity ATHENE. It is co-funded by the European Union (ERC, InterText, 101054961). Views and opinions expressed are, however, those of the author(s) only and do not necessarily reflect those of the European Union or the European Research Council. Neither the European Union nor the granting authority can be held responsible for them. This work benefited from the Federal Ministry of Education and Research (BMBF) projects “PlexPlain” (FKZ 01IS19081) and “XEI” (FKZ 01IS24079B). Further, we gratefully acknowledge the support of Microsoft with a grant for access to the OpenAI GPT models via the Azure cloud (Accelerate Foundation Model Academic Research). Finally, we express our gratitude to the members of the Synthetic RNA Biology Lab for their feedback, the numerous discussions, and their efforts during data annotation.

\bibliography{custom}
\bibliographystyle{acl_natbib}

\newpage

\appendix

\section{Appendix}
\label{sec:appendix}

\subsection{Broader Related Work} \label{asec:extended_rw}
We propose a framework for describing, collecting data, and assisting with structured reasoning during text assessment. Our study relates to a range of application domains of document quality assessment and general work on decomposed explicit human reasoning. Here, we extend the Related Work discussion in Section \ref{sec:rw} to cover the broader related literature.

\paragraph{Explanations and Abductive Reasoning} We frame the LLM experiments on the dataset as an abductive reasoning task, aiming to discover the structured explanation for a given verdict. We contribute to the nascent area of abductive reasoning in NLP \cite{bhagavatula2019abductive}. While prior work primarily focuses on abduction based on formal logic \cite[e.g.][]{young-etal-2022-abductionrules}, we focus on subjective reasoning on a real-world task. Relatedly, \citet{dalal2024inference} evaluate LLMs' ability to infer the best explanation for an observation and provide its reasoning as a free text. In our work, explanations consist of answers to steps in a reasoning graph imposing a consistent explanation structure.
Our work relates to explainability research in NLP \cite[e.g.][]{wiegreffe2021teach,camburu2018snli}. Notably, \citet{dalvi-etal-2021-explaining} evaluate models on constructing reasoning paths (\textit{entailment trees}) to explain science facts. From an explainability perspective, our framework generalizes entailment trees into directed graphs that provide more flexible and comprehensive explanations.

\paragraph{Diagnostic Reasoning} For this study we take inspiration in work on diagnostic reasoning \cite{REITER198757,peng1990abductive}. Structured text assessment can be considered as a form of diagnostic reasoning where a given document is scrutinized to identify the underlying issues in the text that lead to a negative verdict on its quality, akin to a doctor's diagnosis of a patient.

Previous studies in this domain focus on medical \cite{richens2022aiformed,Magnani2023} and engineering \cite{koitz2018applying} problems, primarily using bipartite diagnosis graphs with numerical or categorical variables. In contrast, our framework considers multi-layered graphs, incorporates natural language variables, and leverages LLMs to model relationships between them.

Recently, clinical NLP research has increasingly focused on generating explanations for diagnoses to explicitly evaluate a model's reasoning, which is most related to our study. \citet{10.1001/jamanetworkopen.2024.40969} study human-LLM collaboration in medical diagnosis, investigating whether clinicians’ diagnostic accuracy improves when supported by ChatGPT. In our work, we simulate human-LLM collaboration during diagnosis but limit LLM support to individual reasoning steps. This approach provides more granular insights into the utility of LLMs for various aspects of diagnostic reasoning. \citet{savage2024diagnostic} use LLMs in a zero-shot setting, designing prompts to mimic medical expert reasoning during diagnosis. Complementing this, our study does not focus on prompt design but instead introduces an LLM program architecture informed by expert reasoning processes. Most related to our work are prior studies that use explicit reasoning structures to guide LLMs. For instance, \citet{yang2024drhouse} develop DrHouse, a tool for interactive anamnesis and diagnosis using LLMs augmented with a medical knowledge base. They structure LLM prompts with diagnosis guidelines as binary decision trees to support follow-up questioning and eventual diagnosis. Similarly, \citet{wang2024direct} propose an augmented LLM-program architecture that iteratively generates a chain of diagnostic steps, forming a reasoning trace for the final diagnosis in the style of entailment trees \cite{dalvi-etal-2021-explaining}. In contrast, our study employs directed acyclic graphs to represent reasoning structures. Unlike trees, these graphs allow reasoning steps to recombine multiple prior observations, offering a more flexible framework. Overall, we complement the field of clinical diagnostic reasoning by applying similar techniques to a new domain – diagnosing texts instead of patients – and propose a modeling framework that accommodates the subjectivity and ambiguity inherent to the textual domain, which differs from the medical diagnosis context.

\paragraph{Essay Grading}
Automatic essay grading aims to score student essays' quality and is usually cast as a regression of classification problem \cite{misgna2025survey}. Most related to our study are approaches on feedback generation serving as an actionable explanation for the verdict and that score along a rubric \cite{wang2024beyond}. For example, \citet{han2023fabric} score essays along multiple dimensions serving as an explanation for the overall quality assessment, but they do not consider reasoning structures. \citet{wang2024beyond} design a human-LLM collaboration system for rubric-based scoring, but they do not structure the process. More structured, \citet{fiacco-etal-2022-toward} design a system to annotate essays with rhetorical tree structures serving as a means to identify problematic areas in the essay and produce a quality judgement, but they do not consider complex reasoning over these structures. Our framework connects the paper to the output, so it comes with more expressive explanations and is more fine-grained.

\subsection{Discussion: Paper Assessment as a Explicit Structured Reasoning Process} \label{ssec:paper_diagnosis_discussion}
This Section offers an in-depth discussion of the newly proposed perspective on document and paper assessment as a structured reasoning task. 

\paragraph{Paper Assessment as a \stricta~problem}
We frame paper assessment as a structured reasoning problem in critical text assessment. This framing is plausible and flexible with respect to the partially subjective nature of paper assessment. Specifically, we motivate this framing from the expert discussions and from an analysis of peer review data.

The expert interviews and formative study (see \ref{ssec:interviews}) revealed that assessing a biomedical paper is largely independent of its specific topic. Experts repeatedly investigate the same components, considering the line of argument and underlying research of the paper to determine its quality. Each review differs only in the weight that is assigned to the different aspects, e.g., leaving some out sometimes when irrelevant to the paper type, and how each aspect is determined. This justifies framing paper assessment from the structured reasoning angle because an interconnected set of reasoning variables adequately describes the experts' model of paper quality.

Besides the evidence derived from the expert interviews, we validate the proposed framing within the context of peer review. We analyze $30$ randomly sampled reviews from the NLP domain of the NLPEER corpus \cite{dycke-etal-2023-nlpeer} and manually analyze their lines of argument. This analysis reveals that reviewers often employ what-if-reasoning (e.g. "Without additional experiments covering more models, the expressiveness of the experiments is limited." implying that if the paper had more experiments the reviewer would value the soundness of the paper differently) and that many weaknesses identified by reviewers reoccur across a variety of papers. These observations support the framing of paper assessment as a structured reasoning process and a strong alignment with the causal framework proposed in this paper. 

\paragraph{Advantages of the \stricta~Framing}
Framing paper assessment as a \stricta~problem opens up safe and interpretable human-AI collaboration. In such a scenario, the human expert reads the paper and defines selected variables, and the AI is tasked to fill in the remaining variables automatically or in collaboration with the human. In our experiments, we test different configurations, including providing only the final verdict, and providing full agency to the human on the outcome decision. Since the reasoning structure for the explanation is fixed, the model output becomes easier to interpret.

\paragraph{Relation to Other Forms of Reasoning}
The \stricta~framework allows for defining various reasoning tasks on the underlying workflow depending on the provided data. In the following, we discuss the reasoning task considered during the experiments in Section \ref{sec:llm} -- the task of explaining a given verdict based on the paper text -- in reference to the broader reasoning concepts. 

\begin{figure}[t]
  \centering
  \includegraphics[width=0.8\linewidth]{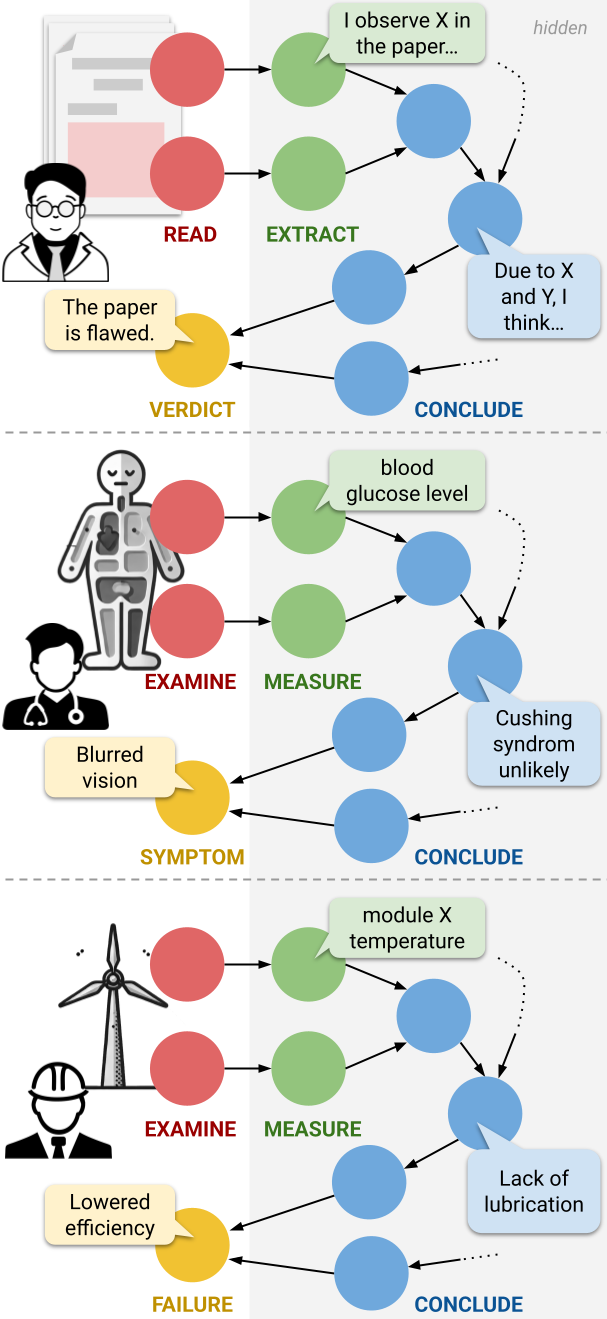}
  \caption{Diagnosis is a general concept used by many professions. Scientists use it during paper assessment (top), physicians use it during medical diagnosis (middle), and engineers apply it to analyze machine failures (bottom). An observed failure behavior (yellow) given the observable system states (red) is explained by assigning the hidden system states a most likely value (grey area).}
  \label{fig:diagnose_comparison}
\end{figure}

The work is inspired by diagnostic reasoning. Diagnosis is a general activity that is not limited to medical diagnosis by physicians. More generally, diagnosis refers to the task of identifying the hidden cause of an observed (malfunctioning) behavior in a system \cite{REITER198757}. For instance, psychotherapists, programmers during debugging of source code, or engineers analyzing machine failures all use diagnostic reasoning \cite[e.g.][]{Mueller2012KnowledgeER,Wong2016ASO,Hameed2009ConditionMA,Magnani2023}.
By the above general definition, paper assessment can be seen as a diagnostic task. Figure \ref{fig:diagnose_comparison} illustrates this in comparison with other diagnostic activities.

Finally, we provide a short discussion on the classification of the considered reasoning task. Abductive reasoning means inferring the most likely explanation for a given set of observations \cite{peirce1955abduction}. Deductive reasoning means deriving a conclusion from a set of premises by applying a sequence of logic rules \cite{leighton2003defining}. We argue that the given problem setting is abductive reasoning rather than deductive. First, unlike deduction, the reasoning process of paper assessment is probabilistic, i.e., reviewers choose the best of all possible explanations for their verdict. The verdict is not derived from a set of formal rules from a set of premises. The subjective nature of the task and the complex interaction of quality criteria make this impossible. Second, in the given task, framing the goal is to find the unobserved intermediate steps that lead to the observed verdict, which aligns precisely with the definition of abduction.

\paragraph{Practical Utility of \stricta-based Assistance}
Conceptually, \stricta~supports effective and interpretable human–AI collaboration in two ways: by enabling explainability and identifying opportunities for human–LLM collaboration. Structured explanations enhance interpretability and explainability of NLP models by augmenting single-shot black-box predictions with intermediate reasoning steps, such as reasoning trees \cite{dalvi-etal-2021-explaining} or multi-hop predictions \cite{jiang-etal-2020-hover}. By design, they allow verification of model outputs, promoting trustworthiness. Structured reasoning also enables systematic analysis of human–LLM collaboration potential (see Section \ref{sec:llm}). It helps identify steps needing assistance, assess LLM competence across reasoning stages, and estimate the feasibility of dynamic task routing between humans and AI.

Beyond these conceptual contributions, we empirically test \stricta’s practical utility during paper assessment. We conduct a small-scale user study with nine postgraduate biomedical students who assess preprints in their field using the dataset generation interface, augmented with GPT-3.5 assistance. Participants could freely engage with the workflow and request AI support at any step. Six completed a post-study survey. On a scale from 1 (fully disagree) to 7 (fully agree), the structured workflow received an average rating of 5.67 for usefulness in paper assessment. The utility of AI assistance during stepwise evaluation was rated 3.8 on a scale from 1 (worst) to 5 (best). These results indicate the potential of \stricta~for effective and satisfactory human–LLM collaboration in text assessment. However, more analysis on the human-LLM interaction dynamics based on large-scale controlled user studies are needed to validate the results of this small-scale study.

\subsubsection{Background on Causality}
\label{ss:causality_background}

Following the definition by \citet{pearl2000models}:
\begin{definition}
    An SCM $\mathcal{M}$ is $(U, V, F, P_\mathcal{M})$, where
    \begin{enumerate}[label=\roman*., noitemsep]
        \item $U$ is a set of exogenous variables determined by factors outside of $\mathcal{M}$. $P_\mathcal{M}$ is a probability function over the domain of $U$.
        \item $V=\{V_1, ..., V_n\}$ is a set of endogenous variables determined by variables in $U \cup V$.
        \item $F=\{f_1, ..., f_n\}$ is a set of functions, the \emph{structural equations}, on the domains of $U_i \cup \textit{pa}_i$ to $V_i$, where $\textit{pa}_i \subseteq V \setminus V_i$, and $U_i \subseteq U$ and $F$ forms a mapping from $U$ to $V$.
    \end{enumerate}
\end{definition}

This means each endogenous (i.e., modeled) variable is given by a structural equation $v_i = f_i(\textit{pa}_i, u_i)$ where $\textit{pa}_i$ is the set of \textit{parent} variables from $V$ and arbitrary exogenous variables $u_i \subseteq U$.
\noindent Exogenous variables are also called \textit{background} variables that add randomness to the otherwise deterministic structural equations. 

\paragraph{Causal Graphs} 
Causal graphs are an intuitive representation of SCMs where the nodes are given by $U \cup V$, and a directed edge indicates that the antecedent is input to the structural equation of the consequent (\textit{child}). We read an edge from $X$ to $Y$ as \textit{X  (may) cause Y}.

\paragraph{Interventions and Counterfactuals}
We notate interventions using the $do$-operator. For an intervention $do(X=x)$, we alter the SCM $\mathcal{M}$ by replacing all occurrences of variable $X$ by $x$ in $F$. We denote the altered model as $\mathcal{M}_{X=x}$. 
\textit{Counterfactuals} pose what-if questions on the SCM. For a query "What value had $Y$ if $X=x$ under the circumstances $u$?" we imagine the causal effect of $X$ on $Y$ for a concrete instance counter to the observed facts. The counterfactual distribution is
$$
P_{\mathcal{M}_{X=x}}(Y=y) = \sum_{\{u | u \in \widehat{U} \land Y_x(u) = y\}} P(u)
$$
where $Y_x(u)$ denotes the potential response of $Y$ under $\mathcal{M}_{X=x}$. Finally, the \textit{average causal effect} (ACE) of an intervention on a binary variable $X$ is the difference of expected values of $Y$ with $\textit{ACE}(X, Y) = \mathbb{E}[Y_{x=1}] - \mathbb{E}[Y_{x=0}]$.

\subsubsection{\stricta~Design Choices}
\label{ss:nldar_alternatives}

\paragraph{Characterizing Reasoning Problems}
For the discussion of alternative formalisms for describing \stricta~ problems we refer to the literature on the nascent field of diagnostic abductive reasoning (DAR) \cite{peng1990abductive} since it also aims to make reasoning explicit and expresses the reasoning as a structured graph.

One very established approach is to model DAR as a set-covering problem \cite{bylander1991computational}. Here, the task is to select a minimal set of hypotheses that explain the observed behavior optimizing for a plausibility metric. Under this framing, the system model consists of a set of independent boolean hypotheses and their relation to the observable behavior. 
As another formalism, propositional logic has been proposed to describe the system model \cite{nordh2008makes}. Here, the hypothesis space is expressed as a set of propositional clauses from which a subset needs to be selected such that it is faithful to the problem context and logically implies the observations. The system model hereby consists of multiple propositional logical clauses and solving the task means finding a satisfiable configuration of hypotheses given the observations. 
Similarly, logic programming models the system as a logic program, posing the behavioral variables as known variables and searching for an optimal answer \cite{lin2002abduction}.

While the above approaches model reasoning as a purely logical and thereby discrete problem, another line of research focuses on probabilistic models of abductive reasoning that explicitly incorporate ambiguity and uncertainty and often support continuous variables. Fuzzy logic models the space of explanations in terms of fuzzy sets, and the solution is discovered by maximizing the probability of the hypothesis set explaining the observable behavior \cite{berner2007clinical,suryanarayanan1995fuzzy}. More prominently, Bayesian networks and structural causal models \cite{richens2020improving}. Here, the underlying causes for observable behavior are expressed in a network of interacting factors. While all formalisms allow for the encoding of a system of components that jointly influence behavioral variables,  logic-based approaches cannot handle ambiguity and uncertainty, whereas statistical approaches can only characterize associational strengths between factors. Only SCMs allow for studying cause and effect relationships  \cite{pearl2000models}, making them strictly more powerful than the other approaches.

\paragraph{Design Choices}
The key challenge in approaching a concrete \stricta~problem or family of \stricta~problems lies in choosing an appropriate structured reasoning model (i.e., the workflow) which is sufficiently expressive yet computational inferences on this model are feasible. 

We opt to describe \stricta~problems using SCMs. On one hand, the probabilistic elements of this formalism can directly model the statistical variance induced by the ambiguity and subjectivity of most document assessment tasks. On the other hand, the ability to pose causal and counterfactual queries on SCMs enables a rich methodology to study the reasoning process of humans and potentially AI models in terms of explanations.

Additionally, for any reasonable text assessment process, the final verdict should depend on the input text of the problem. Therefore, we require the SCM for an \stricta~problem to have three tiers, each depending on the previous tier: the surface text variables, the intermediate reasoning steps, and the final verdict. 

\subsubsection{\stricta~Beyond Paper Assessment}\label{ss:stricta_for_aes}
To exemplify the transfer of \stricta~ to other tasks, we outline the steps necessary to instantiate it for multi-rubric essay scoring. \textbf{Automatic essay scoring} (AES) can enhance teacher consistency and enable large-scale student self-teaching \cite{taghipour-ng-2016-neural}. Beyond predicting an overall score, AES models often provide rubric-specific scores that relate to the final grade \cite{kumar-etal-2022-many}. This task aligns well with \stricta's scope, and we propose a streamlined protocol based on our study. To instantiate \stricta~for essay scoring, we follow these steps (see Section 3.3): (1) design the causal graph, (2) populate it with human data, and (3) perform analysis and partial automation.

To construct the causal graph (1), we first identify the relevant variables, starting with the final verdict as the final score along with a rationale. We model the rubric scores as \textsc{infer} steps that are direct parents to the final verdict. The key challenge lies in defining the information flow from the essay text via \textsc{extract} steps towards the scoring rubrics. Here, we may involve teachers and grading experts to understand their mental workflow. As an example, a rubric score on “narrative quality” \cite{10.1162/tacl_a_00007} is determined by first extracting the exact narrative of the essay and then deriving partial judgements on its consistency and alignment with the essay prompt. As an alternative to the manual definition of the workflow, existing resources of rubric assessments (e.g. \cite{yoo2024dress}) can be used to determine the different factors impacting essay scoring and their relation to the text. To populate the structural causal model with data (2), we replicate the annotation study with human experts based on existing essay corpora, tasking teachers to follow the workflow to grade an essay. Based on our findings, a human-LLM collaborative approach to data generation is promising to speed up this process, and a combination of authentic human data as well as LLM-generated data revised by humans is promising. Finally, we perform an analysis and evaluate LLMs on the task (3). The analysis can be used to quantify the consistency of human ratings (through variability analysis), determine the weighing of rubrics and their localization in the essay text (through average causal effect analysis). Applying LLMs for automatic scoring according to the grading workflow enables students to inspect the specific rationales for the final score and to locate them in the paper. Conceivably, students can use the resulting tool to iteratively improve their paper by producing counterfactual versions that improve their score as a learning exercise. 

Overall, this example illustrates the application of \stricta~ to other critical text assessment scenarios and makes clear the potential for analysis and automation in this domain.

\subsection{Constructing an SCM for Biomedical Paper Assessment}
\subsubsection{Interviews and Formative Study} \label{ssec:interviews}
We report on additional details of the interviews and the formative study used to elicit the workflow from the domain experts. The process was divided into four major steps: before the interviews, the actual interviews, consolidation, and the formative study. For all stages, we worked with two biomedical experts on post-doctoral and professorial levels, respectively. Both have reviewed numerous publications during their careers and have extensive experience in the field.

\paragraph{Before the interviews}
Before the interviews for deriving an SCM, we determine the critical points during text assessment and characteristics of the paper genre to ground all further steps on this information. After interrogating the two experts in a joint session we converge to the following insights on the domain at hand.

First, there are \textbf{two general types of papers} in biomedicine relevant to their work: natural science papers, which investigate natural phenomena, and method papers, which propose a new approach to conducting experiments. Second, biomedical papers follow a \textbf{relatively rigid surface structure} with an introduction including background information, the methods section as a detailed experimental log, the results section reporting the exact measurements, and the discussion section followed by an overall conclusion. The results and discussion section can be intermingled. Third, \textbf{figures and tables play a major role} in the understanding and assessment of a paper. Fourth, \textbf{methodological soundness and consistency of results} are key indicators during text assessment. Finally, several heuristics exist to detect bad experiment design, such as the lack of controls or suspiciously high standard deviations on measurements.

\paragraph{Interviews}
We conducted the interviews with each expert in isolation and oriented towards best practices of interview design by \citet{knott2022interviews}. To structure the interview, we discussed the assessment criteria and process based on reviewing checklists of the field, starting with a general and intuitive description of the assessment process and increasing the level of detail.
For the first interview, we considered the manuscript checklist by \citet{seals2000pr} conceptualized for physiology research. For the second interview, we use the Portland Press review checklist\footnote{\url{https://portlandpress.com/pages/peer_review_checklist} (accessed 08/2024)} for biomedical sciences. We opted for two different checklists to encourage different perspectives during the workflow elicitation and mitigate the impact of the interviewer's biases.

The result of these interviews is two different step-wise manuals on how to iterate through a biomedical paper and assess its methodological soundness. We identify the there types of actions taken during assessment (\textsc{read}, \textsc{extract}, \textsc{infer}) which align between both manuals. While some of the subordinate steps deviate and the researchers put a different emphasize on each of these aspects, the overall assessment approach is very similar. In summary, first, the whole article is skimmed to get an understanding of its contents. Then, the experts check the contextualization with related work and background information in the introduction. Afterward, they skip to the results section, which includes figures and tables. They check the alignment of results and whether the reporting seems plausible and follows best scientific practice. Subsequently, they read the discussion in detail and analyze if the provided interpretations and conclusions align with their own readings of the results and their context in related work. Then, they verify that the method section covers all important methods and that they are described correctly. Finally, they check the conclusion section, abstract, and title to verify that they align with the findings and methodology described in the other sections. Both experts agree on the temporal ordering of these major steps during assessment.

\paragraph{Consolidation}
After the interviews, we consolidate the sequential manuals into one manual covering all the mentioned points by both researchers. We then interpret the temporal ordering and the subordination relationship between major and minor steps as causal dependencies to answer follow-up questions. For instance, to answer a sub-ordinate step, "Is the description of the state-of-the-art complete?" we require the input from the super-ordinate step, "How do the authors describe the state-of-the-art in the field?". 

This results in a workflow of $106$ reasoning steps covering all sections of the paper. We discuss the dependency structure and overall workflow with one of the domain experts and incorporate suggested changes. We use the resulting workflow as input for the formative study.

\paragraph{Formative Study}
We perform the formative study with three PhD students from the same research lab as the two senior experts. Following a similar setup as described in the final annotation study (in Section \ref{sec:data_collection}), we linearized the workflow. We asked them to assess a paper selected by their preference, providing answers for each step. For the formative study, the annotators work without any specific annotation interface and simply append the answers to a text document. We summarize our findings from this formative study in the following.

First, the annotators did not finish the full workflow of $106$ steps within the three hours of the study, suggesting that the workflow should be subselected to keep the annotation practical. Second, the annotators tended to not use the causal dependencies explicitly between the steps, meaning that they answered several steps based on the paper and not referring to their own prior answers. The main obstacle for this lied in the lack of a tailored annotation interface that presents prior answers and enforces the causal structure. Third, annotators need training to get an overview of the workflow graph to avoid redundantly stating similar facts across multiple steps. An easy option to revise prior answers might also be interesting for that purpose.\\

In response to these observations, we designed the final evaluation study setup to collect the dataset underlying the SCM (as described in \ref{sec:data_collection}). More details on the annotation interface can be found in the Appendix \ref{ssec:annotation_study}.

\subsubsection{SCM for Biomedical Paper Assessment}
\label{ssec:biomed_scm}
We give a more extensive overview of the graphical structure, the specific steps, and their prompts in the biomedical paper assessment workflow. 

\begin{figure*}[t]
  \centering
   \includegraphics[width=\linewidth]{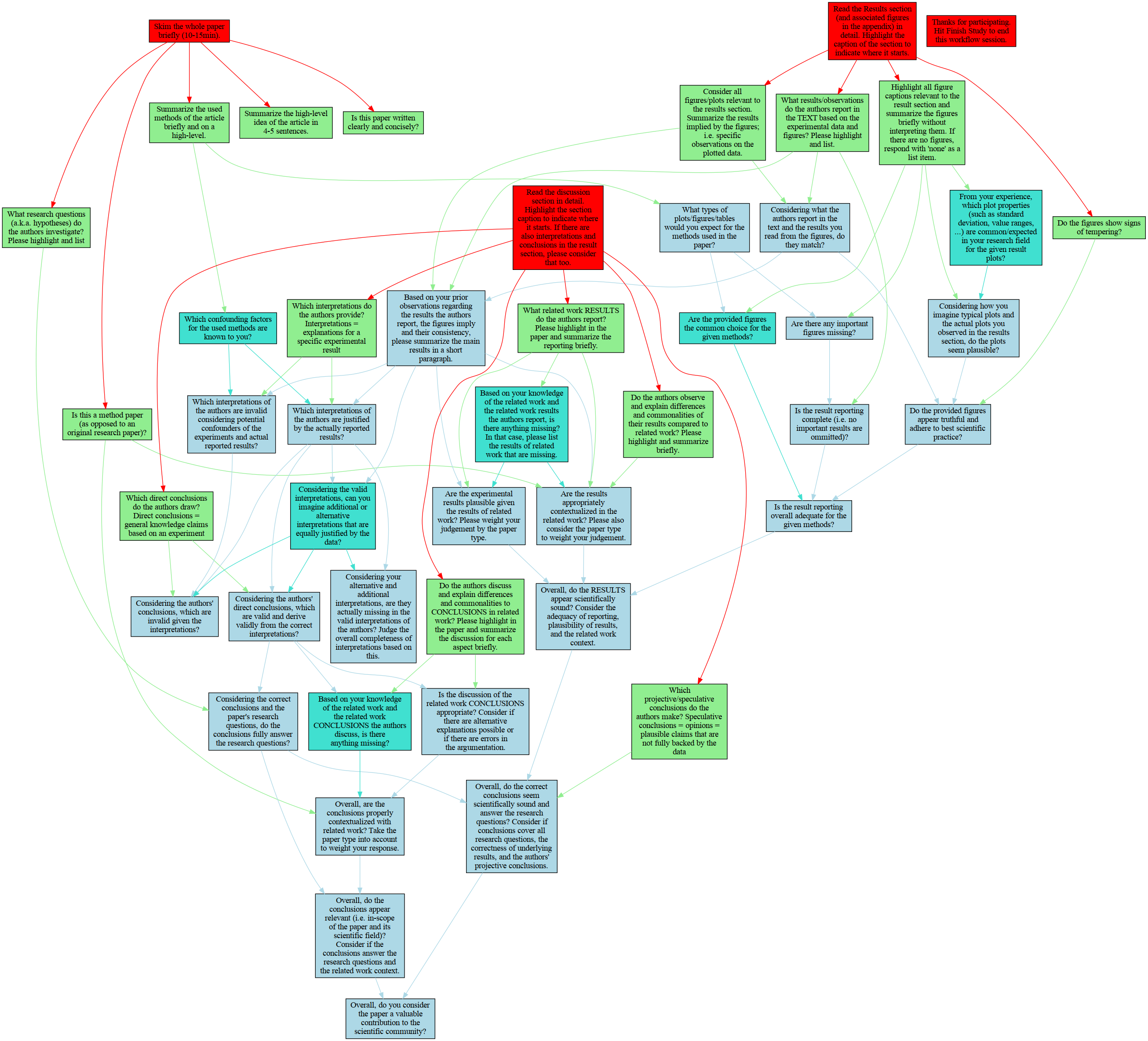}
  \caption{Representation of the full workflow. For each step, the prompt is provided. The color of each node indicates the node type. \textsc{read} nodes are red, \textsc{extract} nodes are green and \textsc{infer} nodes are blue. The turquoise nodes are \textsc{infer} nodes that rely on background information not present in the paper. The edges are colored to match the parent node type and increase the readability.}
  \label{fig:full_workflow}
\end{figure*}

\paragraph{Workflow Steps}
The full causal graph of the biomedical workflow is illustrated in Figure \ref{fig:full_workflow}. Tables \ref{tab:all_workflow_items1} to \ref{tab:all_workflow_items3} list all workflow steps, including their type and prompt. The description and example are omitted for readability. We refer the reader to the associated codebase for the complete workflow step specifications.

\paragraph{Graph Structure}
The graph, as shown in Figure \ref{fig:full_workflow}, is a directed and acyclic graph. The average degree per node lies at $3.57$; the mean number of parents lies at $1.78$; there are $3$ roots in the graph (the \textsc{read}) nodes; there are $4$ terminal nodes of which one is the final verdict, whereas the others are the assessment of clarity and the high-level summary based on the skim. These only exist to ease the annotation and have no meaning for the overall SCM modeling of the DAR problem. 

Upon qualitative inspection of the graph, it becomes clear that the two major sections (results and discussion) form a dense network of descendants, while there are only a few connections between the nodes belonging to the different sections. Only the nodes closer to the final verdict share parents belonging to different sections. This allows the graph to be divided into $5$ layers of nodes, where each layer contains only nodes that can be assessed in parallel according to topological ordering. We make use of this property for the batch-wise computation of the LLM programs during our experiments described in Section \ref{sec:llm}.

\paragraph{Boolean Workflow Version}
For our causal analysis of impact factors on decision making (see Section \ref{sec:analysis}), we convert the full text-based workflow into one that consists only of the boolean nodes to estimate the structural equations from data. 

\begin{figure*}[t]
  \centering
   \includegraphics[width=\linewidth]{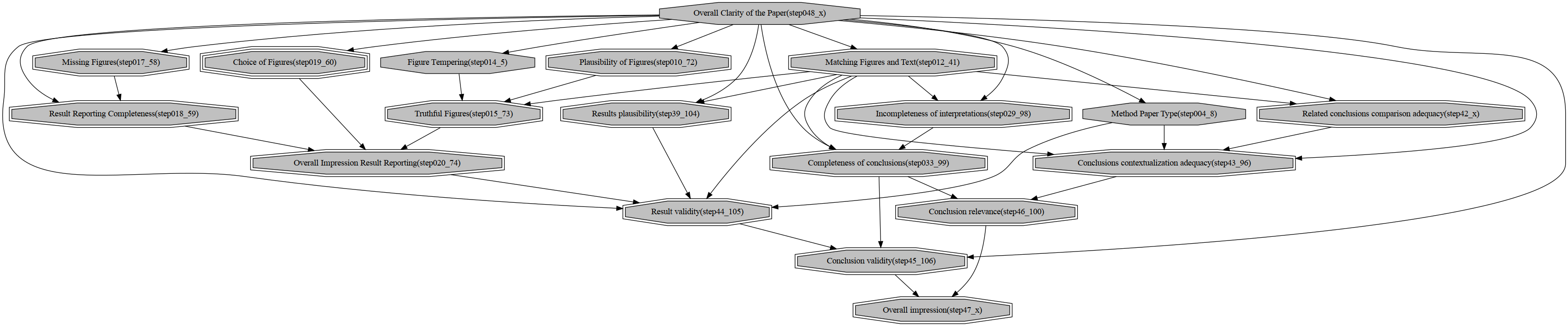}
  \caption{Representation of the boolean workflow. For each step, the name and ID are provided.}
  \label{fig:bin_workflow}
\end{figure*}

Figure \ref{fig:bin_workflow} shows the causal graph of the resulting boolean workflow. This graph maintains the causal ancestry relationships of the original graph. More formally, the output graph maintains the transitive closure of the original graph on the subset of selected nodes.
To arrive at this, we substitute all text-only nodes (such as \textsc{extract} nodes) with an edge, which means that we basically "skip" this step but maintain the causal relationship from their parents to their children. For nodes that would be substituted and lie at the end of a path, we drop the nodes. This is the case for read nodes. However, as all read nodes depend on the paper, we add an additional joint dependency of all children of read nodes to the node assessing the clarity of the paper. We make the assumption that the read nodes of different sections are independent except for the clarity of the paper acting as a confounder. 

\begin{table*}[t]
    \centering
    \renewcommand{\arraystretch}{1.2}
    \resizebox{0.99\linewidth}{!}{
    \begin{tabular}{P{2cm}P{3cm}P{1.5cm}P{8.5cm}}
         \toprule
         \textbf{Id} & \textbf{Name} & \textbf{Type} & \textbf{Prompt} \\
        \midrule
         step001.42 & Skim & read & Skim the whole paper briefly (10-15min). \\
step002.44 & High-level Idea & extract & Summarize the high-level idea of the article in 4-5 sentences. \\
step034.50 & Research Questions & extract & What research questions (a.k.a. hypotheses) do the authors investigate? Please highlight and list \\
step003.43 & High-level Methods & extract & Summarize the used methods of the article briefly and on a high-level. \\
step004.8 & Method Paper Type & extract & Is this a method paper (as opposed to an original research paper)? \\
step048.x & Overall Clarity of the Paper & extract & Is this paper written clearly and concisely? \\
step006.4 & Results Section & read & Read the Results section (and associated figures in the appendix) in detail. Highlight the caption of the section to indicate where it starts. \\
step007.7 & Figures & extract & Highlight all figure captions relevant to the result section and summarize the figures briefly without interpreting them. If there are no figures, respond with 'none' as a list step. \\
step016.25 & Expected Figures for Methods & infer & What types of plots/figures/tables would you expect for the methods used in the paper? \\
step019.60 & Choice of Figures & infer knowledge & Are the provided figures the common choice for the given methods? \\
step017.58 & Missing Figures & infer & Are there any important figures missing? \\
step008.6 & Results in Figures & extract & Consider all figures/plots relevant to the results section. Summarize the results implied by the figures; i.e. specific observations on the plotted data. \\
step009.71 & Typical Figure Properties & infer knowledge & From your experience, which plot properties (such as standard deviation, value ranges, ...) are common/expected in your research field for the given result plots? \\
step010.72 & Plausibility of Figures & infer & Considering how you imagine typical plots and the actual plots you observed in the results section, do the plots seem plausible? \\
step011.28 & Results in the Text & extract & What results/observations do the authors report in the TEXT based on the experimental data and figures? Please highlight and list. \\
step012.41 & Matching Figures and Text & infer & Considering what the authors report in the text and the results you read from the figures, do they match? \\
         \bottomrule
    \end{tabular}
    }
    \caption{The first steps of the workflow with ID, type, and prompt.}
    \label{tab:all_workflow_items1}
\end{table*}

\begin{table*}[t]
    \centering
    \renewcommand{\arraystretch}{1.2}
    \resizebox{0.99\linewidth}{!}{
    \begin{tabular}{P{2cm}P{3cm}P{1.5cm}P{8.5cm}}
         \toprule
         \textbf{Id} & \textbf{Name} & \textbf{Type} & \textbf{Prompt} \\
        \midrule
        step013.29 & Summary of Results & infer & Based on your prior observations regarding the results the authors report, the figures imply and their consistency, please summarize the main results in a short paragraph. \\
step014.5 & Figure Tempering & extract & Do the figures show signs of tempering? \\
step015.73 & Truthful Figures & infer & Do the provided figures appear truthful and adhere to best scientific practice? \\
step018.59 & Result Reporting Completeness & infer & Is the result reporting complete (i.e. no important results are ommitted)? \\
step020.74 & Overall Impression Result Reporting & infer & Is the result reporting overall adequate for the given methods? \\
         step021.12 & Discussion Section & read & Read the discussion section in detail. Highlight the section caption to indicate where it starts. If there are also interpretations and conclusions in the result section, please consider that too. \\
step022.20 & Interpretations & extract & Which interpretations do the authors provide? Interpretations = explanations for a specific experimental result \\
step023.19 & Direct conclusions & extract & Which direct conclusions do the authors draw? Direct conclusions = general knowledge claims based on an experiment \\
step024.18 & Projective conclusions & extract & Which projective/speculative conclusions do the authors make? Speculative conclusions = opinions = plausible claims that are not fully backed by the data \\
step025.26 & Method confounders & infer knowledge & Which confounding factors for the used methods are known to you? \\
step026.30 & INVALID interpretations & infer & Which interpretations of the authors are invalid considering potential confounders of the experiments and actual reported results? \\
step027.33 & VALID interpretations & infer & Which interpretations of the authors are justified by the actually reported results? \\
step028.34 & ALTERNATIVE interpretations & infer knowledge & Considering the valid interpretations, can you imagine additional or alternative interpretations that are equally justified by the data? \\
step029.98 & Incompleteness of interpretations & infer & Considering your alternative and additional interpretations, are they actually missing in the valid interpretations of the authors? Judge the overall completeness of interpretations based on this. \\
step030.35 & VALID conclusions & infer & Considering the authors' direct conclusions, which are valid and derive validly from the correct interpretations?  \\
step031.31 & INVALID conclusions & infer & Considering the authors' conclusions, which are invalid given the interpretations? \\
         \bottomrule
    \end{tabular}
    }
    \caption{The intermediate steps of the workflow with ID, type, and prompt.}
    \label{tab:all_workflow_items2}
\end{table*}

\begin{table*}[t]
    \centering
    \renewcommand{\arraystretch}{1.2}
    \resizebox{0.99\linewidth}{!}{
    \begin{tabular}{P{2cm}P{3cm}P{1.5cm}P{8.5cm}}
         \toprule
         \textbf{Id} & \textbf{Name} & \textbf{Type} & \textbf{Prompt} \\
        \midrule
step033.99 & Completeness of conclusions & infer & Considering the correct conclusions and the paper's research questions, do the conclusions fully answer the research questions? \\
step035.17 & Related work results & extract & What related work RESULTS do the authors report? Please highlight in the paper and summarize the reporting briefly. \\
step36.14 & Related work results differences & extract & Do the authors observe and explain differences and commonalities of their results compared to related work? Please highlight and summarize briefly. \\
step37.102 & Related work results missing & infer knowledge & Based on your knowledge of the related work and the related work results the authors report, is there anything missing? In that case, please list the results of related work that are missing. \\
step38.103 & Results contextualization & infer & Are the results appropriately contextualized in the related work? Please also consider the paper type to weight your judgement. \\
step39.104 & Results plausibility & infer & Are the experimental results plausible given the results of related work? Please weight your judgement by the paper type. \\
step44.105 & Result validity & infer & Overall, do the RESULTS appear scientifically sound? Consider the adequacy of reporting, plausibility of results, and the related work context. \\
step40.13 & Related conclusions discussion & extract & Do the authors discuss and explain differences and commonalities to CONCLUSIONS in related work? Please highlight in the paper and summarize the discussion for each aspect briefly. \\
step42.x & Related conclusions comparison adequacy & infer & Is the discussion of the related work CONCLUSIONS appropriate? Consider if there are alternative explanations possible or if there are errors in the argumentation. \\
step41.15 & Related conclusions missing & infer knowledge & Based on your knowledge of the related work and the related work CONCLUSIONS the authors discuss, is there anything missing? \\
step43.96 & Conclusions contextualization adequacy & infer & Overall, are the conclusions properly contextualized with related work? Take the paper type into account to weight your response. \\
step45.106 & Conclusion validity & infer & Overall, do the correct conclusions seem scientifically sound and answer the research questions? Consider if conclusions cover all research questions, the correctness of underlying results, and the authors' projective conclusions. \\
step46.100 & Conclusion relevance & infer & Overall, do the conclusions appear relevant (i.e. in-scope of the paper and its scientific field)? Consider if the conclusions answer the research questions and the related work context. \\
step47.x & Overall impression & infer & Overall, do you consider the paper a valuable contribution to the scientific community? \\
         \bottomrule
    \end{tabular}
    }
    \caption{The final steps of the workflow with ID, type, and prompt.}
    \label{tab:all_workflow_items3}
\end{table*} 

\newpage
\subsection{Data Collection}
\subsubsection{Underlying Paper Dataset} \label{ssec:underlying_papers}
We crawl the underlying papers from bioRxiv and ensure a high sample diversity along relevant dimensions. Here, we outline the details of this process.

\paragraph{Data Acquisition}
We retrieve the index of all papers on bioRxiv using a custom crawler based on the recent paper list\footnote{\url{https://www.biorxiv.org/content/early/recent}, accessed 08/2024}. We deduplicated the list, resulting in roughly $100$k in preprint URLs. We then retrieve all paper meta-data, including the title, authors, corresponding author institution, type of study, version, category tag, publication information, license, abstract, and the number of versions of each article, through the bioRxiv API. We augment each entry by the country and world region of the authoring institution\footnote{We match the institutions via the publicly available university domain mapping \url{https://github.com/Hipo/university-domains-list}} and the paper length estimated from the number of PDF pages.\\

Next, we filter all preprints to include only those with a license that permits redistribution (CC0, CC-BY, CC-BY-NC, CC0-NG), that belong to any of the relevant categories in $\{$biochemistry, bioengineering, synthetic biology, microbiology, molecular biology, bioinformatics$\}$, and whose abstracts contain at least one of the terms $\{$RNA, regulatory RNA, synthetic, RNA circuit, biosensor, riboswitch, aptamer$\}$.\footnote{Our expert annotators provided this list.} Finally, we filter by publication date (published between 06/2021 and 06/2023) to ensure recency. This results in $1252$ eligible articles. Then, we estimate the covered topics using LDA \cite{NIPS2010_71f6278d} on the abstracts of the remaining preprints. 

\begin{figure}
    \centering
    \includegraphics[width=1\linewidth]{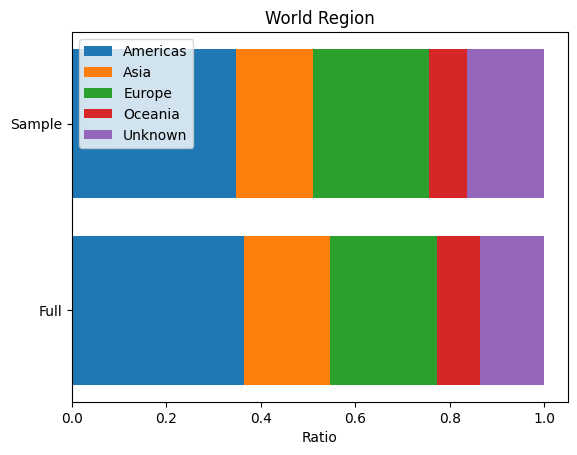}
    \caption{The ratio of each world region in the full paper dataset and the subsample of $22$ papers included in the annotation study.}
    \label{fig:world_region_distribution}
\end{figure}

\begin{figure}
    \centering
    \includegraphics[width=1\linewidth]{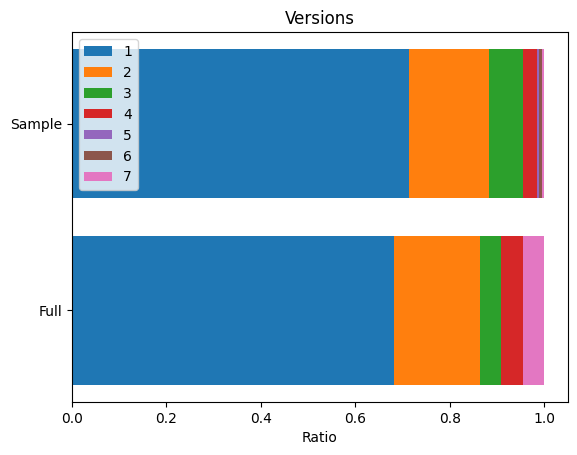}
    \caption{The ratio of versions in the full paper dataset and the subsample of $22$ papers included in the annotation study.}
    \label{fig:versions}
\end{figure}

\begin{figure}
    \centering
    \includegraphics[width=1\linewidth]{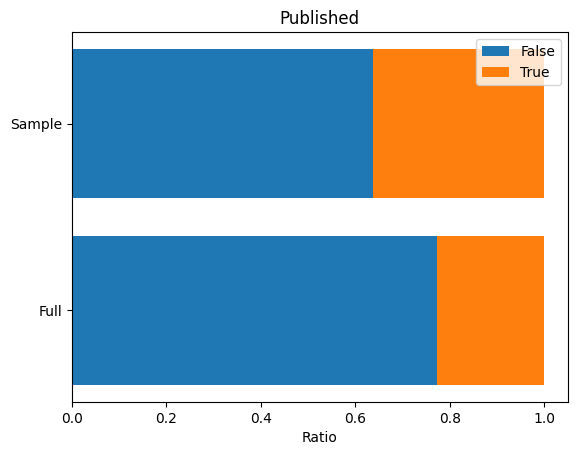}
    \caption{The ratio of publication status in the full paper dataset and the subsample of $22$ papers included in the annotation study.}
    \label{fig:published}
\end{figure}

\paragraph{Sample Diversity}
Based on this dataset, we aim to select a representative subset along the dimensions of the world region of the authoring institution, the number of versions that exist of the preprint, and the publication status. The latter two criteria should ensure that there are preprints of varying quality in the samples, whereas the first should ensure that the papers come from diverse institutions with different publication records and writing styles. For this purpose, we take a stratified sample of $200$, which is then manually inspected and filtered by one expert annotator, resulting in $57$ eligible papers, of which $22$ were annotated.
Figures \ref{fig:world_region_distribution} to \ref{fig:published} show the sample diversity of the final dataset along the key dimensions as compared to the initial stratified sample of $200$ papers. While observing a light skew towards accepted papers, all world regions, versions, and publication statuses are present in similar proportions as in the full sample. All papers are written in English which is the lingua franca across science. 

\paragraph{Data Format}
For all resulting papers, we retrieve the PDF and the XML from bioRxiv. To complement the XML with figures and tables, which are represented as images, we reverse engineer their URLs and retrieve those as JPG files. We parse the XML together with the images and store them in Intertext Graph Format (ITG)\footnote{\url{https://github.com/UKPLab/intertext-graph/tree/main}}. 

\subsubsection{Annotation Study Details} \label{ssec:annotation_study}

\paragraph{Overview}
We employed the following guidelines and annotation interface to ensure maximal consistency among annotators and to encourage step-wise causal thinking. As another key quality measure, we repeatedly engaged with individual annotators and the whole annotator group to calibrate and provide feedback on their annotations.

\paragraph{Guidelines} The guidelines consisted of the following text extended by a sequence of images explaining the annotation interface in detail.\\

{ \small
A “reviewing workflow” is a formal way of reasoning about the quality of a paper. A workflow consists of a sequence of questions that build on top of each other. The workflow was developed with experts of the field and while it might not resemble your personal internal mindmap during peer review, it is a good way of discussing an article’s quality.

Some of the questions in the workflow simply extract information from the article or your background knowledge, others infer or combine quality judgements based on earlier answers. you will encounter the following questions/tasks in a workflow:
read – here you read a section or part of the paper. 
extract – based on the section you read (in an earlier question) you try to extract certain information from that section. Usually this means you will need to highlight some information in the paper and write a summary.
infer – based on previous answers to questions, you make a judgment on the question (usually a yes/no question) by combining the previous answers in the right way.

Your task in the annotation study is to strictly follow the reviewing workflow as you are guided through its questions step-by-step. While this seems unnatural at first, this makes sure that all participants follow the same reasoning, while still answering each question naturally and from their subjective point of view. During the annotation there are several things to keep in mind, which are explained in more detail below. Most importantly, you should try to annotate each question based on precisely the input that is provided to you during the workflow; do not answer questions in isolation and by a vague general feeling of the paper.

\underline{Reviewing Workflow Example} \\
In the following you will be guided through a dummy workflow to make you more familiar with the annotation interface and the concept of workflows. Before you dive into CARE, please have a look at the right example for the underlying logic of workflow. When you do the annotation and in this mini-tutorial this graph of dependent questions is converted into a questionnaire.

Please click on the following link to go through the tutorial. If you should have problems with stepping through the workflow, please refer to the section Annotation Interface or talk to the study coordinators. We recommend Firefox for using CARE. 

\underline{Terminology} \\
Some terminology re-occurs in many questions of the workflow and there is some room for misinterpretation. Please briefly consider the following terms and “our” definition for them.

\begin{itemize}
    \item Research question (aka hypothesis): A research question or hypothesis = a claim (possibly phrased as a question) on the system under investigation which the authors try to verify or falsify; example: 'is brain cancer caused by this DNA sequence?'
    \item Method (aka approach): The methods or approach of a paper = the (standardized) tools, technologies, and actions the authors undertake to approach the research question; example: modular cloning, Gibson assembly, ...
    \item Result: The results of a paper = the data measured/generated during the experiments of the paper. The results comprise more or less the raw measurements, possibly aggregated by statistics, but not interpreted.
    \item Interpretation: An interpretation = based on results (raw data) the authors provide an interpretation, i.e. an explanation or conclusion within just the scope of this specific experiment run, but without claiming correctness of the statement for the whole natural science world; example: the experiments show an upward trend in variable X.
    \item Conclusion (aka finding): A conclusion = based on the interpretations of experimental results and the validity of the experimental setup, the authors generalize a claim from the specific experiment to the natural science world; example: variable X and variable Y correlate.
    \item Method Paper Type:  A method paper introduces a new experimental setup / technology / tool / technique to solve a practical research problem; e.g. discovering a new RNA lever to enable follow-up research or designing an algorithm to measure quantity X. The central research question usually is “Can we develop a new tool for task X that has properties Y?”. The method of such a paper comprises the newly proposed tool/setup/technique. The results cover the measurements on desired properties Y on task X (e.g. accuracy) of the tool. The interpretations analyze the significance of these properties (compared to related work) and the conclusions are usually claims about whether the new tool is an improvement to existing tools along certain dimensions and for which scientific endeavors it should be applied.
    \item Original Research Paper Type: An original research paper investigates a research question on natural science phenomena; e.g. checking if two molecules interact with each other or searching for an explanation for certain behavior in cells. The central research question can be diverse; they are usually a causal question (“does A cause B?” / “why does X behave as observed?” / etc.). The method of such a paper comprises the experimental setup based on any prior tools/approaches and  how they are combined. The results cover measurements on the observed variables and the interpretations analyze these results to form specific claims about the experiments. The conclusion usually puts the interpretations into perspective and makes claims about the underlying natural science phenomenon.
\end{itemize}

\underline{Principles}
\begin{itemize}
\item English Language – please answer the questions only in English.
\item Fluency – In general, answer all questions with fluent text. 
\item Explanations – For yes/no answers, please always first provide your assessment (yes / no), but then also add an explanation that describes how you arrive at this answer and contextualizes/constraints your assessment. This is especially important when you cannot give a clear yes/no response. E.g.: Q: “Are there outdated references?” (good) A: "Yes. There are some references to methods that are clearly outdated: they are more than 5 years old. However, the references to the most relevant work on the same issue are recent, so I judge the references to be overall sufficiently new.” (bad, trivial explanation) A: “Yes. Because the references are not outdated.”
\item Good reasoning (what-if) – on infer questions, where you combine and weigh prior answers to respond to the new question, you should ask yourself, what input answers would turn my judgment positive/negative. Likewise, you should check that all information in your answer really comes only from the inputs.
\item Uncertain Flag – You can mark an answer as uncertain (flag icon), if you really don’t feel able to answer the question. However, use this option very sparingly!
\item Highlighting – For the extractive questions you should try to highlight whole sentences and not on sub-sentence level. Ideally, for each aspect you extract you can narrow down the highlight to at most 4 consecutive sentences – avoid selecting whole paragraphs, let alone pages. If you need to highlight across two pages (not possible in CARE), please simply make two highlights.
\item Reading Tasks – if a reading task points you to a certain paper section, do consider if this section is present as such or if you should peek into others to answer the questions; e.g. some discussion might appear  in results section.
\item Comprehensive Answers – Remember that all your answers to questions will form the input to later steps in the workflow. This means your answers should be comprehensive to allow all follow-up judgements. If you notice at a later question that your earlier answer is insufficient, do not hesitate to jump back to that earlier question and flesh it out. With some practice in using the workflow, this will become less frequent. 
\item Don’t be too picky  and don’t be too nice. The purpose of the reviewing workflow is to perform a mostly neutral assessment of the paper at hand. It is not about “destroying” the paper (even if you disagree with the goals/ideas/… of it) and you shouldn’t naively trust everything the authors say or get distracted by the authors' good or bad writing style.
\end{itemize}

\underline{Examples} \\
Q: What is the pool of related work you know?

Bad: I somewhat remember there are some other works that deal with that phenomenon. → informal, vague

Better: There is a list of related works that deals with the same phenomenon: X, Y, Z. → formal, precise

Better: I am not familiar with the detailed related work of the field, however works X, Y, Z appear similar to the scope of this work. → formal, objective \\

Q: Are there references missing? (based on 2 and 3)

Bad: Yes. → no explanation, no context

Bad: No, it really sounds like they cover all bases. → informal, not derived from 2 + 3, but a general feeling on the paper directly

Better: Yes. While they reference the most important works (X, Y), they omit Z, which also investigates <subject>.  → contextualized, explained \\

Q: Are there outdated references? (based on x, y)

Bad: Yes → no explanation, no context

Bad: Yes, because there are outdated references → trivial explanation

Better: Yes. The references to the methodology are all older than 10 years. In the given fast-paced area of research such references are generally outdated. → contextualized, explained

}

\begin{figure*}
    \centering
    \includegraphics[width=1\linewidth]{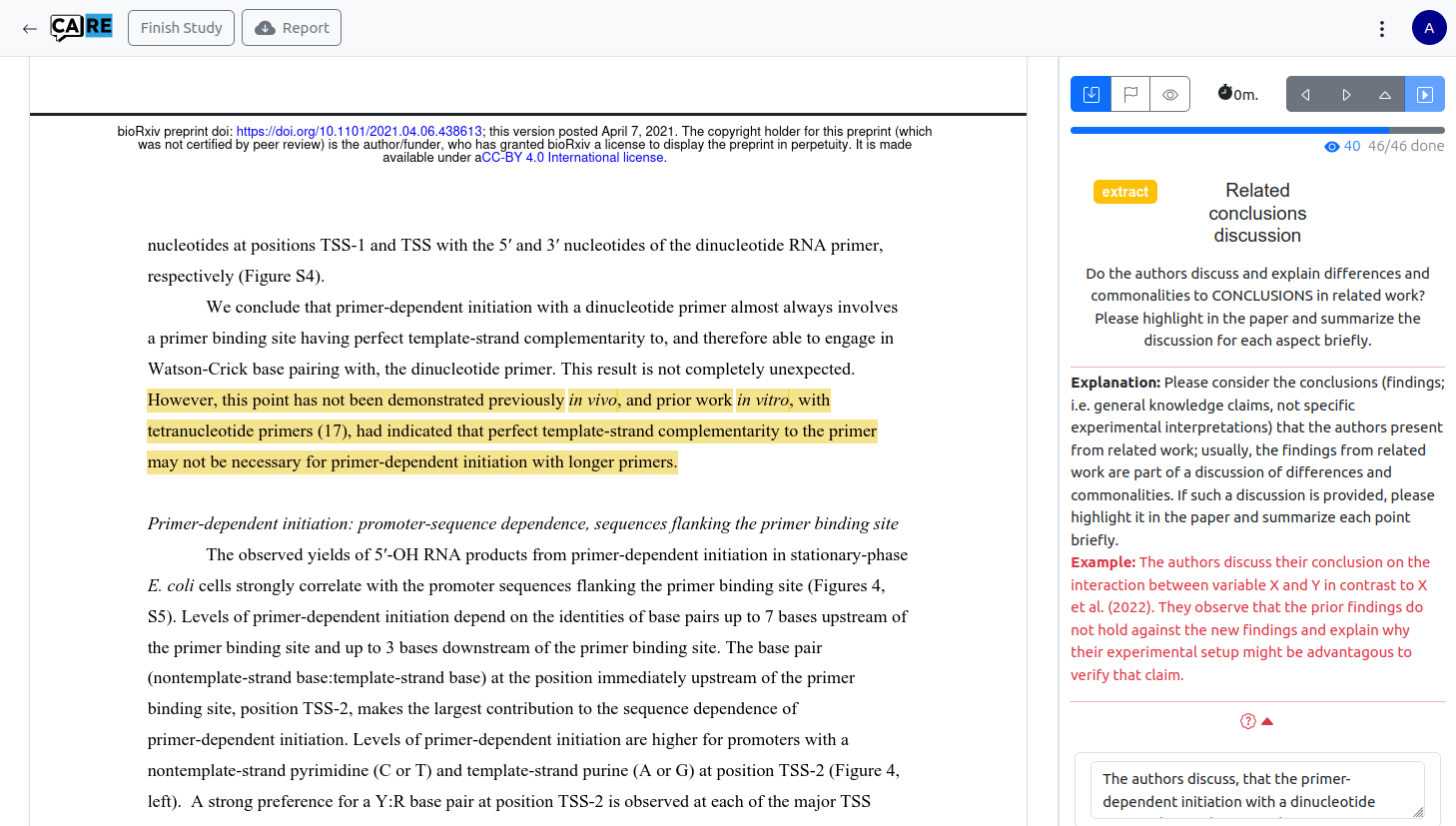}
    \caption{The annotation interface for an extract step. The paper with highlights is on the left; on the right, the user receives the task for the given step. They can answer in a structure or free-text form, depending on the question. Annotators can skip through the steps, flag them, and jump to prior steps as needed.}
    \label{fig:care_ai1}
\end{figure*}

\begin{figure*}
    \centering
    \includegraphics[width=1\linewidth]{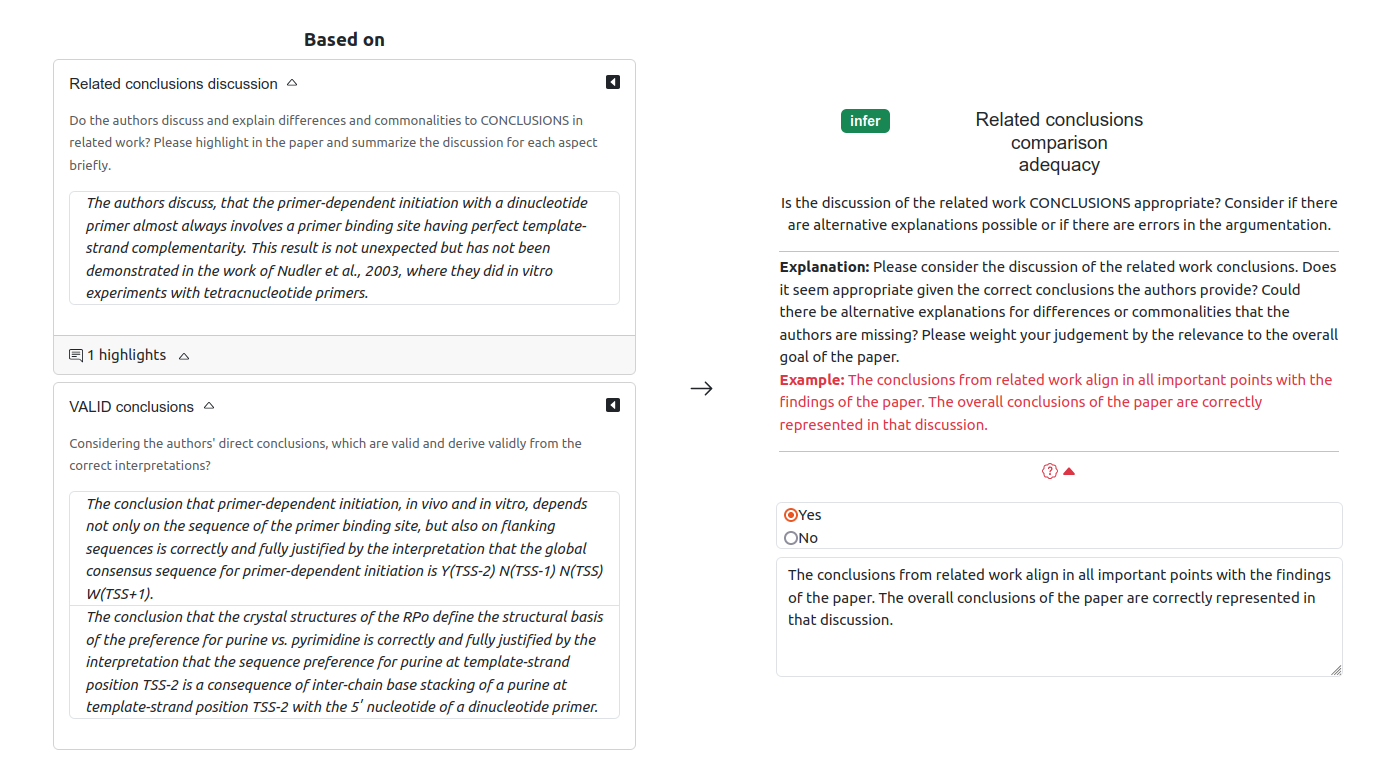}
    \caption{The annotation interface for inference steps. Here, a modal opens above the paper to block the view and encourage annotators to concentrate only on the provided inputs of prior steps (on the left) to write their answers.}
    \label{fig:care_ai2}
\end{figure*}

\paragraph{Annotation Interface}
Screenshots of the annotation interface are shown in Figures \ref{fig:care_ai1} and \ref{fig:care_ai2}. On the right, the annotators can control how they step through the workflow. They can jump back to earlier steps if needed. In the view for \textsc{infer} items, annotators cannot see the paper and have to base their answer on their prior answers.

\paragraph{Annotator Demographics}
We report selected annotator statistics including demographics for reproducibility. Due to the relatively small pool of annotators, we report the annotators of the \textsc{Junior} and \textsc{Senior} study jointly and omit detailed characteristics that could violate their anonymity. The identities of all annotators are kept secret and we assign random pseudonyms in the dataset to avoid deanonymization.

\begin{figure}
    \centering
    \includegraphics[width=1\linewidth]{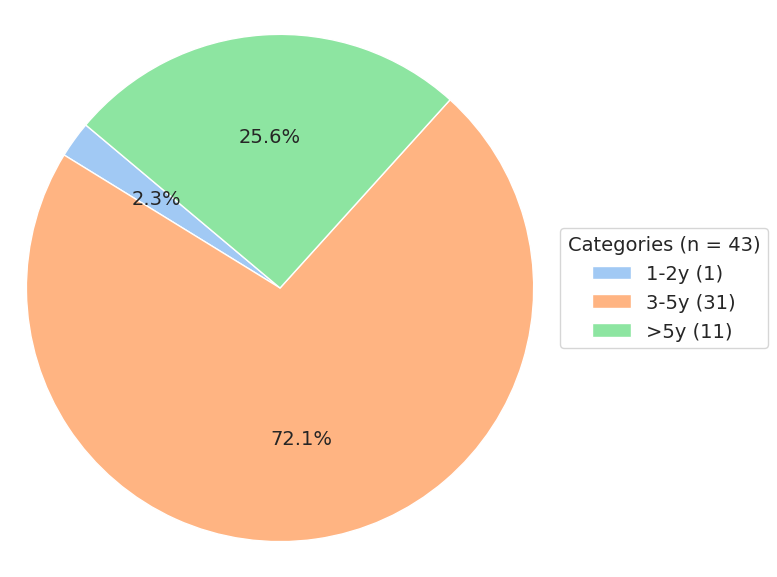}
    \caption{Self-reported number of years concerned with biomedical studies.}
    \label{fig:years}
\end{figure}

\begin{figure}
    \centering
    \includegraphics[width=1\linewidth]{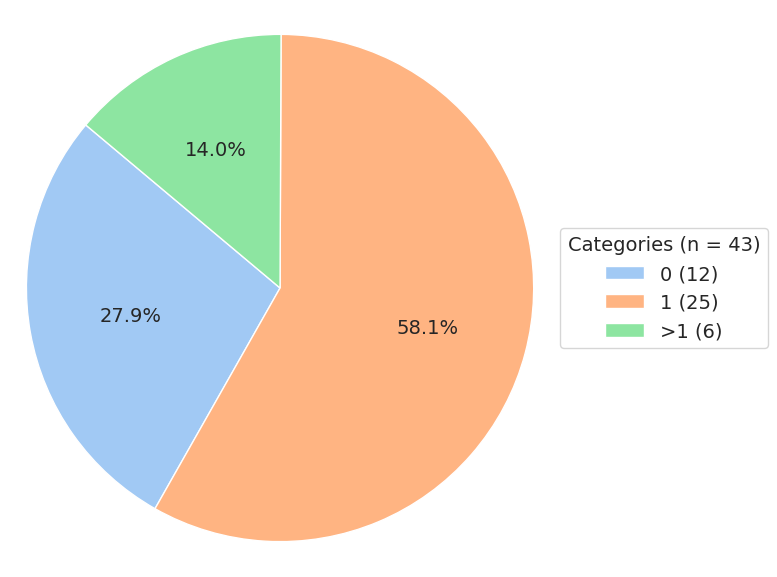}
    \caption{Self-reported number of papers read per week on average.}
    \label{fig:papers_per_week}
\end{figure}

The distribution of genders is nearly balanced between male and female; diverse was never self-reported. Nearly all annotators are from the age group $20-29$. Figure \ref{fig:years} reports the number of years each annotator is engaged in biomedical research in any form. Figure \ref{fig:papers_per_week} shows the distribution of biomedical papers read per week in the whole pool. Several of the \textsc{Junior} annotators do not regularly read biomedical papers, but several consume them occasionally. The \textsc{Senior} annotators all consume papers regularly. Nearly all annotators speak English as a second language.

\paragraph{Dataset Post-processing}
We pseudoynmize all data and bring the annotations into a unified format. Because the \textsc{Junior} annotators have less experience with respect to academic review and publishing, we filter this dataset by a set of quality heuristics to ensure high quality. As a proxy for quality, we manually inspect the answers of each annotator. We discard annotators' answers that provide a drastically different summary of the article compared to other annotators and the abstract. We further discard annotations by annotators that provide consistently shorter answers than the other annotators. Finally, we discard annotations which contain multiple trivial explanations ("A is because A is.") or provide non-English answers. Overall, we filter highly restrictively to ensure high quality of the dataset. We discard annotations by $20$ of the \textsc{Junior} annotators.

\paragraph{Dataset Size and Generalizability}
We take several measures to ensure that the findigns from our analysis (Section \ref{sec:analysis}) and LLM experiments (Section \ref{sec:llm}) on the dataset are general and representative. As detailed beforehand, we ensure representativity of the paper sample by sampling biomedical preprints across multiple dimensions. The dataset captures roughly $4000$ reasoning steps, with three to five expert responses per paper, contributed by roughly $40$ experts in over seven hundred hours of work. We deem this amount of data sufficient for our  analysis.

To investigate whether the dataset size is sufficient to make general findings on LLM performance, we perform a statistical analysis. Specifically, we investigate whether adding more papers would affect our experimental findings. To examine this empirically, we investigate a proxy: the impact of a single paper on LLM evaluation. If removing or adding one paper to the dataset does not significantly alter the results, the dataset can be considered adequate and the findings generalizable. 

We test this hypothesis using statistical analysis. We focus on BERT-F1 scores as the evaluation metric. The null hypothesis states that the mean performance on the full dataset equals the mean performance on subsets excluding one paper. Only if the dataset is of sufficient size, leaving out a single paper will not induce high variance in the results. For each leave-one-paper-out subsampled dataset, we compute the mean performance and compare it to the overall mean performance computed on all papers using a single-sample t-test. To account for multiple models and therefore multiple testing, we apply Bonferroni correction. At a significance level of $0.05$ (Bonferroni-corrected to $0.0163$), the null hypothesis is not rejected; p-values exceed 0.98 across all models (specifically $0.9872$, $0.9852$, and $0.9898$).
These results indicate no significant effect on performance evaluation when omitting a paper from the dataset. Thus, the dataset size is sufficient for the objectives of this study since the results are not sensitive to adding or removing a paper. 

\subsection{Transfer Study} \label{ssec:transfer}
To validate the generalizability of the framework, we test the transfer of the framework to case of paper assessment in natural language processing with distinct requirements. We report the detailed results here.

We employ in total $33$ PhD students specialized in NLP research to conduct a paper assessment based on the adapted workflow. This resulted in $9$ full executions (answering every item) and $24$ partial executions of the workflow. We received nearly $1000$ answers to individual reasoning steps on top of six NLP papers. To validate the workflow for the domain, we conducted an optional survey among the participants for which we received ten responses.

Rated on a scale from 1 (strongly disagree) to 7 (strongly agree), 60\% of respondents assigned a score of at least 5 (above neutral) to the alignment with personal assessment practices. In free-form feedback, some of the participants noted that some steps appear to discuss overlapping paper aspects. Others noted that the provided schema worked very well for them. 
Overall, the transfer study results indicate that adapting the proposed approach to paper assessment is feasible with reasonable effort.

\subsection{Data Insights}
\subsubsection{Detailed Analysis of Answer Variability} \label{ssec:variability_details}

We complement the analysis of answer variability in the main paper with a detailed analysis of variability throughout the graph.

\paragraph{Does variability depend on the areas of the paper that are analyzed?}
When considering the semantic variability along the workflow graph, we see a weak negative correlation (Kendall's $\tau = -0.36$) between the position of the step in the linearized workflow and the average semantic similarity. Here, the number of preceding steps seems to play a role (Kendall's $\tau = -0.15$), but also the scope of the steps focusing either on high-level information (mean $0.60$ cosine similarity), results ($0.63$), or interpretations and conclusions ($0.48$) in the paper. This is plausible given that the discussion section of a paper gives rise to more subjective assessments than the raw results. All in all, the variability depends on the elements of the paper that are discussed.

\paragraph{Does variability propagate through the workflow graph?}
Considering the relatively high semantic alignment between annotators on extraction tasks (mean semantic similarity of $0.60$), a lot of the answer variability has to be the consequence of deviating opinions during intermediate reasoning steps despite relying on the same evidence. To discover the impact of deviating opinions, we investigate the propagation effect of answer variability along paths of the workflow graph and for individual annotators.

We focus on answers that are substantially different from the ones of the other annotators for a step. We define an answer as substantially disagreeing (\textit{dis}) if the mean semantic similarity to the other annotators lies one standard deviation below the average similarity across all papers for that step. This means, if an annotator shows exceptionally high dissimilarity with the other annotators we flag this answer as a substantial disagreement. Under this assumption, roughly $21\%$ of all answers show substantial disagreements. We can then compute the probability of propagating disagreement as
$P(\textit{dis}|\geq 1 \textit{ parent with dis}) = 0.72$. As expected, substantial disagreement propagates to the subsequent questions with a $72\%$ chance as compared to random disagreement with only $21\%$.

\subsubsection{Structural Equation Approximation from Data} \label{ssec:factors}

We report on the boolean decision SCM for analysis. Specifically, we outline the goodness of fit for the resulting SCM trained on the human-generated dataset.

\paragraph{Boolean SCM Representation}
The model is represented as an invertible structural causal model within the do-why library\footnote{\url{https://www.pywhy.org/dowhy/v0.11.1/}}. The root nodes of the graph are estimated as empirical distributions, and the other nodes are assumed to be sampled with additive noise. To learn the mechanism of a node, we use Gaussian process classifiers. These assumptions allow us to estimate the structural equations from data and determine the noise variables for counterfactual queries. 

\paragraph{Goodness of Fit}
We rely on the \texttt{evaluate\_causal\_model} function of the do-why library to estimate how well the learned SCM approximates the observed human decisions to measure its validity.

The graph structure of the SCM aligns well with the data, as there are no other permutations of the DAG inside its Markov equivalence class. The average KL divergence between the distribution implied by the learned structural equations lies at $0.18$ which shows an overall close fit to the data. The invertibility assumption of the nodes is not rejected for any of the nodes except the root node \texttt{step4}, which determines if the paper is of the type \textit{method paper}. Notably, the Continuous Ranked Probability Score is low for all nodes (below $0.31$) except for \texttt{step4} at $0.39$. While this hints at the fact that the model might not perfectly reflect this factor, the overall model fits the observed data well.

\subsubsection{Implementation Details}
We report on the implementation details for the analysis of the SCM.

To determine the semantic similarity we use SBERT \cite{reimers2019sbert} provided by the \texttt{sentence-transformers} library (version 2.7.0). We use the embedding model "sentence-transformers/all-mpnet-base-v2" trained on MPNET. For the similarity by part-of-speech tags and lemmas, we rely on spacy's\footnote{\url{https://spacy.io/} \texttt{spacy-v3.6.1} and \texttt{scispacy-v0.5.3}} language processing module specialized for the scientific domain \texttt{en\_core\_sci\_lg}.

\subsection{LLM Assistance Experiments}
We report on hyperparameters and all details used for the different experiments. In total, the experiments with commercial models invoked roughly $200$ USD of API inference costs. The open models were run on three A100 GPUs.

\subsubsection{Experimental Configurations} \label{ssec:llm_configs}
We report on the detailed LLM versions, hyperparameters, and prompts for the three experimental scenarios (automation, human-LLM simulation, and ablations on error propagation). Additionally, we report on the evaluation metric versions used.

\paragraph{Open LLM Configurations}
We use the langchain\footnote{\url{https://www.langchain.com/}, \texttt{langchain-v0.2.8}, \texttt{langchain-openai-v0.1.19}} and huggingface transformers\footnote{\url{https://huggingface.co/}, \texttt{transformers-v4.35.2}} libraries to load the open LLMs for inference.

We use the instruction-tuned LLama-3 model with 8B parameters and the extend 32k context window\footnote{\url{https://huggingface.co/NurtureAI/Meta-Llama-3-8B-Instruct-32k}} to fit full papers into the input for \textsc{extract} nodes. For Mixtral, we use the mixture of 8 experts with 8B parameters instruction-tuned\footnote{\url{https://huggingface.co/mistralai/Mixtral-8x7B-Instruct-v0.1}}. We also tested other models including Llama-3 with 70B parameters but the context size was insufficient to fit the relevant inputs.

We run all open models with the generation configuration specified in Table \ref{tab:configuration_llms}. For Llama we add the necessary special tokens for termination.

\begin{table}
    \centering
    \begin{tabular}{cc}
        \toprule
        \textbf{parameter} & \textbf{value} \\
        \midrule
         \textbf{max new tokens}& 2048\\
         \textbf{temperature} & 0.0001\\
         \textbf{top p}& 0.95\\
         \textbf{do sample} & True\\
         \textbf{repetition penalty} & 1.15 \\
         \bottomrule
    \end{tabular}
    \caption{LLM Generation Configuration}
    \label{tab:configuration_llms}
\end{table}

\paragraph{Closed LLM Configurations}
For the closed LLMs we use the AZURE OpenAI API\footnote{\url{https://azure.microsoft.com/en-us/products/ai-services/openai-service}}. We use the models GPT-3.5-turbo with a 16k context window \footnote{\url{https://platform.openai.com/docs/models/gpt-3-5-turbo}, \texttt{gpt-35-turbo-0613-16k}} and GPT-4o\footnote{\url{https://platform.openai.com/docs/models/gpt-4o}, \texttt{gpt-4o}} with default generation parameters.

\paragraph{Prompts and Prompt Selection}
We design two prompts for each model, one for the \textsc{extract} steps and one for the \textsc{infer} steps. Table \ref{tab:llama3} shows the prompts for LLama-3, Table \ref{tab:mixtral} for Mixtral, Table \ref{tab:gpt35} for GPT-3.5-turbo, and Table \ref{tab:gpt4} for GPT-4o. The template parameters are substituted before inference: the \textit{parent} parameter is replaced by the paper text or prior answers in the graph; the \textit{task} parameter is replaced by the prompt for that step in the workflow; the \textit{description} is replaced by the detailed task description of that step in the workflow; the \textit{format} specifies a json object matching the expected output type (e.g. including a boolean field for boolean questions); \textit{example} is substituted by the same example that annotators saw during annotation but in a suitable json format. For feedback and refinement, we use the prompts specified in Table~\ref{tab:feedback_prompt} and in Table~\ref{tab:refine_prompt} for all models. Here, the \textit{true verdict} is replaced by the decision and associated explanation of the task's verdict. The \textit{graph} is specified using incident encoding on integer numbered nodes; i.e. node connections are provided in the format "node x is parent of node y". The \textit{title} takes the paper's title and the answers it the list of initial answers as JSON objects including a restatement of the task prompt and dependencies for each step.

We first tune the prompts by manual inspection until the output format and answers appear plausible. We then tune them manually for an optimal BERT-Score on the development set of the data. We do not perform an extensive search for the best prompt templates; conceptually, we could tune a prompt for each step individually. This is infeasible within the scope of this work.

\begin{table}
    \centering
    \renewcommand{\arraystretch}{1.3}
    \small
    \begin{tabular}{rP{4.8cm}}
         \toprule
         \textbf{Type} & \textbf{Prompt} \\
         \midrule
         \textsc{extract} & "system": "You are a biomedical researcher assessing the quality of a research article."
"human": "This is the research article text:
\{parents\}

Based on the article text, answer the following question:
\{task\}
More specifically, this means: \{description\}

Keep your answer concise. Return the answer as a JSON object in the following format:
\{format\}

Here's an example output for the given question:
"\{example\}".
"\\
         \midrule
         \textsc{infer} & "system": "You are a biomedical researcher assessing the quality of a research article."
"human": "You already gathered the following insights on the paper by answering a sequence of prior questions:

Your insights:
\{parents\}

Based on these prior insights, answer the following question:
\{task\}
More specifically, this means: \{description\}

Keep your answer concise. Return the answer as a JSON object in the following format:
\{format\}

Here's an example output for the given question:
"\{example\}".
"\\
         \bottomrule
    \end{tabular}
    \caption{GPT-3.5-turbo prompts in chat format. \{...\} mark template parameters.}
    \label{tab:gpt35}
\end{table}

\begin{table}
    \centering
    \renewcommand{\arraystretch}{1.3}
    \small
    \begin{tabular}{rP{4.8cm}}
         \toprule
         \textbf{Type} & \textbf{Prompt} \\
         \midrule
         \textsc{extract} & "system": "You are a biomedical researcher assessing the quality of a research article."
"human": "This is the research article text:
\{parents\}

Based on the article text and figure descriptions, answer the following question:
\{task\}
More specifically, this means: \{description\}

Keep your answer concise. Return the answer as a JSON object in the following format:
\{format\}

Here's an example output for the given question:
"\{example\}".
"
"human": \{img\_caption\}
"human": \{img\_base64\} \\
         \midrule
         \textsc{infer} & "system": "You are a biomedical researcher assessing the quality of a research article."
"human": "You already gathered the following insights on the paper by answering a sequence of prior questions:

Your insights:
\{parents\}

Based on these prior insights, answer the following question:
\{task\}
More specifically, this means: \{description\}

Keep your answer concise. Return the answer as a JSON object in the following format:
\{format\}

Here's an example output for the given question:
"\{example\}".
"\\
         \bottomrule
    \end{tabular}
    \caption{GPT-4o prompts in chat format. \{...\} mark template parameters. The \textsc{infer} prompt is identical to the one of GPT-3.5-turbo. For \textsc{extract} steps, the image caption and data messages are repeated for each figure and table relevant to multimodal steps.}
    \label{tab:gpt4}
\end{table}

\begin{table}
    \centering
    \renewcommand{\arraystretch}{1.3}
    \small
    \begin{tabular}{rP{4.8cm}}
         \toprule
         \textbf{Type} & \textbf{Prompt} \\
         \midrule
         \textsc{extract} & "system": "You are a biomedical researcher assessing the quality of a research article."
"human": "The research article text:
\{parents\}

Based on the article text, answer the following question:
\{task\}
More specifically, this means: \{description\}

Keep your answer concise. Return the answer as a JSON object in the following format:
\{format\}

Here's an example output for the given question:
"\{example\}"." \\
         \midrule
         \textsc{infer} & "system": "You are a biomedical researcher assessing the quality of a research article."
"user":"You already gathered the following insights on the paper by answering a sequence of prior questions:

Your insights:
\{parents\}

Based on these prior insights, answer the following question:
\{task\}
More specifically, this means: \{description\}

Keep your answer concise. Return the answer as a JSON object in the following format:
\{format\}

Here's an example output for the given question:
"\{example\}"." \\
         \bottomrule
    \end{tabular}
    \caption{Llama3 prompts in chat format. \{...\} mark template parameters.}
    \label{tab:llama3}
\end{table}

\begin{table}
    \centering
    \renewcommand{\arraystretch}{1.3}
    \small
    \begin{tabular}{rP{4.8cm}}
         \toprule
         \textbf{Type} & \textbf{Prompt} \\
         \midrule
         \textsc{extract} & "human": "You are a biomedical researcher assessing the quality of a research article. The research article text:

\{parents\}

Based on the article text, answer the following question:
\{task\}
More specifically, this means: \{description\}

Keep your answer concise. Return the answer as a JSON object in the following format:
\{format\}

Here's an example output for the given question:
"\{example\}"." \\
         \midrule
         \textsc{infer} & "human": "You are a biomedical researcher assessing the quality of a research article. You already gathered the following insights on the paper by answering a sequence of prior questions:

Your insights:
\{parents\}

Based on these prior insights, answer the following question:
\{task\}
More specifically, this means: \{description\}

Keep your answer concise. Return the answer as a JSON object in the following format:
\{format\}

Here's an example output for the given question:
"\{example\}"." \\
         \bottomrule
    \end{tabular}
    \caption{Mixtral prompts in chat format. \{...\} mark template parameters.}
    \label{tab:mixtral}
\end{table}

\begin{table}
    \centering
    \renewcommand{\arraystretch}{1.3}
    \small
    \begin{tabular}{P{7.2cm}}
         \toprule
         \textbf{Feedback Prompt} \\
         \midrule
         "system": "You are a specialized AI assistant focused on biomedical research quality assessment and
    scientific paper evaluation. You have expertise in analyzing research methodology, study design,
    and the logical relationships between different quality metrics.", "human": "Task: Evaluate and provide feedback on a set of dependent answers assessing the quality of the biomedical article "\{title\}".

    Context:
    - You are reviewing answers to a sequence of quality assessment questions
    - The questions form a directed acyclic graph (DAG) of dependencies
    - Each answer should only depend on its parent nodes in the graph
    - The final goal is to validate or correct the assessment to match the known true verdict *exactly* (i.e. not only the decision, but the explanation for it)

    Input Structure:
    1. Article Title: "\{title\}"
    2. Set of Answers:
       \{answers\}
       Format expected for answers:
       \{\{
            "question\_id": \{\{
                 "node": "question\_id",
                 "question": "question",
                 "question\_type": "type of question",
                 "answer": json object with answer,
                 "parent\_nodes": [list of parents]
            \}\}
       \}\}
    3. Dependency Graph: \{graph\}
    4. True Final Verdict: \{true\_verdict\}

    Required Analysis:
    1. Consistency Check:
       - Verify that each answer only uses information from its parent nodes
       - Identify any circular reasoning or skipped dependencies

    2. Logical Flow Analysis:
       - Trace the path from initial answers to final verdict
       - Identify breaks in the logical chain
       - Flag contradictions between connected answers

    3. Verdict Alignment:
       - Compare the current final verdict with the true verdict
       - If different, identify the minimal set of answers that need revision

    Output Requirements:
    1. Start your response with "Feedback:"
    2. For each suggested revision:
       - Identify the specific answer(s) needing change
       - Explain why the change is needed
       - Describe how the change affects dependent answers
       - Provide specific suggestions for improvement
    3. Prioritize changes that:
       - Maintain the dependency structure
       - Require minimal modifications
       - Lead to the correct final verdict

    Example Answer Format:
    Feedback:
    1. Critical Issues:
       - [List of major inconsistencies]
    2. Required Revisions:
       - Answer X: [Current] → [Suggested revision]
         Reason: [Explanation]
         Impact: [Effects on dependent answers]
    3. Optional Improvements:
       - [Additional suggestions for clarity/completeness]"
    \end{tabular}
    \caption{Feedback prompt used by all models during self-refinement. \{...\} Mark template parameters. For Mixtral the system prompt is presented as a regular human prompt. The graph is provided using incident encoding using integer indexes to describe the nodes.}
    \label{tab:feedback_prompt}
\end{table}

\begin{table}
    \centering
    \renewcommand{\arraystretch}{1.3}
    \small
    \begin{tabular}{P{7.2cm}}
         \toprule
         \textbf{Refine Prompt} \\
         \midrule
         "system": "You are a helpful AI assistant specializing in biomedical research and academic paper quality assessment.
You are skilled at analyzing complex dependency relationships and ensuring logical consistency in multi-step evaluations.", "human": "Task: Revise a set of interdependent answers evaluating the quality of a biomedical article based on provided feedback and a known final verdict.

Input Format:
- title: The title of the biomedical article being evaluated: "{title}"
- answers: JSON object containing question IDs and their current answers
\{answers\}
- graph: Directed acyclic graph in adjacency list format showing question dependencies
\{graph\}
- true\_verdict: The validated final quality assessment of the article:
\{true\_verdict\}
- feedback: List of specific comments on the current answers:
\{feedback\}

Dependencies:
- The graph shows how questions relate: an edge from X→Y means Y's answer should only depend on X and other parent nodes
- All revised answers must maintain logical consistency with their dependent nodes
- The final verdict must be *exactly* supported by the chain of revised answers; i.e. the decision and the explanation must be supported
- The new final verdict should use the true verdict verbatim

Constraints:
1. Maintain the original graph structure
2. Make minimal necessary changes to align with feedback
3. Ensure all answers support the true\_verdict
4. Preserve the logical flow from parent to child nodes
5. Keep the original answer format unless feedback specifically suggests changes

Expected Output Format:
\{\{
    "question\_id": \{\{
         "node": "question\_id",
         "question": "question",
         "question\_type": "type of question",
         "answer": json object with answer,
         "parent\_nodes": [list of parents]
    \}\}
\}\}

Please maintain this structured format in your revised answers and ensure all changes are traceable to either the feedback or the need to align with the true verdict. Return only the JSON object containing the corrected answers with the true verdict.
    \end{tabular}
    \caption{Refine prompt used by all models during self-refinement. \{...\} Mark template parameters. For Mixtral the system prompt is presented as a regular human prompt. The graph is provided using incident encoding using integer indexes to describe the nodes.}
    \label{tab:refine_prompt}
\end{table}

\paragraph{Evaluation Metrics}

For evaluation, we use the BERT-F1-Score from the huggingface evaluate library\footnote{\url{https://huggingface.co/docs/evaluate/en/index} \texttt{evaluate-v0.4.2}} and choose \texttt{distilbert-base-uncased} as the base model. We use the huggingface version of the TRUE score\footnote{\url{https://huggingface.co/google/t5_xxl_true_nli_mixture}} and choose \texttt{google/t5\_xxl\_true\_nli\_mixture} as the model. We parse the first character as the output of the model and assume a default score of zero if the answer cannot be parsed.
For SummaC we use the author provided version\footnote{\url{https://github.com/tingofurro/summac} \texttt{summac-v0.0.4}}.

\subsubsection{Extended Results}
\label{ss:extended_results}
In this section we report on more results mentioned in the main paper.

\paragraph{Additional Ablations and Metrics}

Table \ref{tab:performance_ablations} shows the full results including additional ablations and all metrics. Besides measuring the performance of models in the self-refinement paradigm, we test them by executing them just as LLM programs without any form of backtracking to ensure consistency to the final verdict. Furthermore, we report the F1-score both considering the individual annotators answer as ground truth and when aggregating them to a majority label. 

Finally, we also consider the factual alignment of humans and LLMs to the paper text for \textsc{extract} steps in terms of the SummaC score (Table \ref{tab:performance_ablations} shows it as SummaC$_\text{p}$). We compute it using the paper text as a reference to determine if the responses summarize the paper well. However, \textsc{extract} steps do not describe a classical summarization tasks, instead they aim to reduce the amount of presented information to a specific aspect such as the interpretations in the paper. Seemingly, humans compress information more than LLMs since the SummaC score is notably lower than for the LLMs. This supports our findings from the manual inspection that LLMs tend to create an exhaustive list of facts as opposed to humans that focus on facts relevant to their assessment.

\begin{table*}[t]
    \centering
    \small
    \begin{tabular}{cccccccc}
    \toprule
        & \textbf{BERT-F1} $\uparrow$& \textbf{SummaC} $\uparrow$ & \textbf{TRUE} $\uparrow$ & \textbf{SummaC$_\text{p}$} $\uparrow$ &\textbf{F1-maj} $\uparrow$& \textbf{F1-ind} $\uparrow$\\
        \midrule
        human$^{*\dagger}$ & $\textbf{0.799}_{\pm.06}$ & $-0.151_{\pm.30}$ & $0.151_{\pm.27}$ & $-0.235_{\pm0.34}$ & $\textbf{0.859}$ & $\textbf{0.801}$ \vspace{2mm} \\      
        
        Llama3 Prg$^{\dagger}$ & $0.752_{\pm0.10}$ & $-0.274_{\pm0.36}$ &  $0.098_{\pm0.30}$ & $-0.138_{\pm0.274}$ & $0.151$ & $0.170$\\
        Mixtral Prg$^{\dagger}$ & $0.761_{\pm0.09}$ & $\textbf{-0.149}_{\pm0.26}$ & $0.120_{\pm0.32}$& $-0.109_{\pm0.22}$ & $0.553$ & $0.559$ \\
        GPT3.5t Prg$^{\dagger}$ & $0.759_{\pm0.10}$ & $-0.178_{\pm0.35}$ & $\textbf{0.163}_{\pm0.37}$ & $\textbf{-0.102}_{\pm0.262}$  & $0.529$ & $0.531$ \\
        GPT4o Prg$^\dagger$ & $0.780_{\pm0.07}$ & $-0.186_{\pm0.30}$ &  $0.139_{\pm0.35}$ & $-0.138_{\pm0.28}$ & $0.718$ & $0.720$ \\
        
        \midrule
        Llama3 Prg$^{\beta}$ & $0.752_{\pm.10}$ & $-0.274_{\pm.36}$ & $0.098_{\pm.30}$ & & $0.151$ & $0.170$  \\
        Mixtral Prg$^{\beta}$ & $0.761_{\pm.09}$ & $\textbf{-0.149}_{\pm.26}$ & $0.120_{\pm.32}$ & & $0.553$ & $0.559$  \\
        GPT3.5t  Prg$^{\beta}$ & $0.750_{\pm.11}$ & $-0.165_{\pm.33}$ & $0.159_{\pm.37}$ & & $0.513$ & $0.512$  \\
        GPT4o  Prg$^{\beta}$ & $0.779_{\pm.07}$ & $-0.196_{\pm.29}$ & $0.133_{\pm.34}$ &  & $0.696$ & $0.699$ \\
        \midrule  
        \midrule
        majority$^{io}$ &  & & & & & $0.854$\\
        human$^{*io}$ & $0.799_{\pm.06}$ & $-0.158_{\pm.29}$ & $0.150_{\pm.27}$ & & & $0.801$ \vspace{2mm} \\
        Llama3$^{io}$ & $0.786_{\pm.07}$ & $-0.141_{\pm.30}$ & $0.145_{\pm.35}$ & & & $0.657$ \\
        Mixtral$^{io}$ & $0.794_{\pm.07}$ & $\textbf{-0.077}_{\pm.27}$ & $0.161_{\pm.37}$ & & & $0.822$ \\
        GPT3.5t$^{io}$ & $\textbf{0.805}_{\pm.07}$ & $-0.125_{\pm.34}$ & $\textbf{0.214}_{\pm.41}$ & &  & $0.789$ \\
        GPT4o$^{io}$ & $0.795_{\pm.07}$ & $-0.094_{\pm.28}$ & $0.194_{\pm.40}$ & & & $\textbf{0.876}$ \\
        \midrule
        \midrule
        GPT4o$^\S$ & $0.776_{\pm0.07}$ & $-0.154_{\pm0.28}$ & $0.188_{\pm0.39}$ & & $0.843$ & $0.828$  \\
        GPT4o$^\gamma$ & $0.775_{\pm.07}$ & $-0.156_{\pm.28}$ & $0.187_{\pm.39}$ & & $0.842$ & $0.827$ \\
        \bottomrule
    \end{tabular}
    \caption{Performance under varying conditions. $\cdot^\dagger =$ LLMs run as a program \textit{with} feedback and revise loop that considers the final verdict. $\cdot^\beta =$ LLMs run as a program \textit{without} feedback or refinement neglecting the true final verdict. $\cdot^{io} = $ LLMs receive the human input for each step. GPT4o$^\S$ receives prompts and the full paper text, answers them sequentially without using the graph structure, and applies feedback and refine on top of those. GPT4o$^\gamma$ receives prompts and the full paper text sequentially and does \textit{not} use feedback and refinement. We average over eligible steps and papers, with the standard deviation in subscript. For all metrics, the higher, the better. The human baselines are lower bound estimates of human performance because all answers are compared despite different verdicts. F1-maj refers to the F1 score comparing decisions to the majority voted decisions per node and F1-ind refers to the F1 score comparing decisions to the individual annotator's answers as groundtruth. . SummaC$_\text{p}$ computes SummaC against the paper text computed only on \textsc{extract} steps. All scores except SummaC lie in the range $[0, 1]$. SummaC lies in $[-1,1]$.}
    \label{tab:performance_ablations}
\end{table*}

\paragraph{Error Propagation in the Graph}
The stark performance gap between LLMs run as a program and human-LLM joint performance suggests error propagation as a factor. To examine, we discard the graph structure and run GPT-4o independently for each step, using the full paper as input and applying the same self-refinement loop. Results, shown in the last line of Table \ref{tab:performance}, reveal improved performance compared to GPT-4o operating as a program, though still below its performance with human oversight. This suggests that errors on earlier steps in the workflow deprive later steps of information or provide incorrect information for prediction.

\subsubsection{Human Evaluation} \label{ss:human_eval}
The use of automatic evaluation in our study is a necessity, as human evaluation of all system outputs would be prohibitively expensive and prevent iterative development. While the automatic metrics used in this work are widely adopted in the community, they still remain only imperfect approximators of the real performance perceived by the users. To gain deeper insights into the relationship between our automatic metrics and human judgements for the tasks at hand, we conducted a pilot study with human evaluators. %

\paragraph{Annotator Pool}
We hire $7$ human annotators with biomedical experience through the Prolific platform\footnote{\url{https://app.prolific.com/}}. All annotators are paid at least $10.40$\pounds~per hour and their payment approved regardless of their annotation performance. We require the annotators to have at least a post-graduate degree in biomedical or biochemical sciences to ensure relevant expertise. All annotators self-report being proficient in English; half of them are native speakers. All annotators indicate that they read biomedical papers regularly. We provide two abstracts of papers included in the study asking annotators to verify their expertise with respect to the specific field (synthetic biology) before taking up the task.

\paragraph{Setup}
The objective of the human evaluation is to assess how closely the LLM-generated responses and other human answers align with the true human answer for a given workflow step. To streamline this process, we select $8$ out of the $45$ workflow steps, focusing on early-stage steps (\textit{extract} steps) and critical boolean decision steps identified by their average causal effect on the final outcome (see Section \ref{sec:analysis}). The evaluation is divided into two parts based on these step types.
In the first part, annotators determine which facts extracted by the LLM are implied by the original human answer. A dedicated interface displays the predicted and the ground-truth facts. Annotators assess whether each predicted fact is implied by one, multiple, or none of the original facts. We call this task the \textit{matching} task. In the second part, annotators evaluate the alignment of two explanations for boolean decisions. They assess whether the facts in the explanations contradict or align and whether similar points are raised, ignoring the decision outcome itself. We call this task the \textit{explanation alignment} task.

For each workflow item, we compare the true human answer with two other human answers, the response from Mixtral (using self-refinement and paper text with human figure descriptions), and the response from GPT-4o (using self-refinement and paper text with figures as images).  We choose these models because they are the top-performing LLMs. For all $8$ workflow steps, we select $3$ papers and pick the response of one annotator from the original study as the ground truth.
For each item, annotators view one combination of these outputs alongside the true human answer to assess alignment. Annotation items are randomly sampled and presented in random order. Annotators first evaluate \textsc{extract} steps, then boolean \textsc{infer} steps, using provided guidelines and examples.
We screen the annotator answers considering the time of annotation per item and selected items with a known correct solution. 

\paragraph{Resulting Data}
The annotation study results in $40$ annotation items of which $14$ items across the $8$ workflow steps have at least two annotations. $7$ annotators participated in the study; we filter out one annotator based on the described screening.

We measure the level of inter-annotator agreement using Krippendorff's $\alpha$ with categorical labels for each of the tasks. For the \textit{matching} task, the IAA lies at $\alpha = 0.412$ computed on all pairwise comparisons of facts, for \textit{explanation alignment} at $\alpha = 0.444$. This moderate level of agreement is similar to other tasks aiming to identify links between two texts (e.g. implicit linking in \citet{kuznetsov2022revise}) and point to the difficulty of the task similar to annotation scenarios in argumentation mining \cite{lawrence2020argument,iskender2021argument}. Extensive annotator training, paper context, and a rectification protocol are appropriate measures to foster consistency \cite{klie2024analyzing}. We leave this to future work because we deem the given level of consistency sufficient to estimate the relative performance differences on the task.

\paragraph{Evaluation Results}
We aggregate the \textit{matching} annotations to compute the average overlap between the facts of the predicted answer and the original human answer. We compute the Jaccard metric considering matched items as shared items for each annotator and then average over all. For \textit{explanation alignement} we take the majority vote by all annotators on an item and compute the average amount of aligned explanations. Table \ref{tab:human_eval_results} shows the results. The results closely resemble the findings based on the automatic metrics with Mixtral and GPT-4o performing nearly on-par and the human conservative baseline lagging slightly behind. GPT-4o's performance is better on boolean decision steps than Mixtral but worse during extraction of facts.

Finally, we estimate how well the human evaluation scores align with the automatic metrics. Specifically, we compute the Pearson correlation to determine whether there is a linear relationship with the automatic metrics. Table \ref{tab:human_eval_correlation} shows the correlation coefficients for the two evaluation scenarios. BERT-F1 overall correlates moderately with human judgement in both scenarios. As a general purpose text similarity metric, BERT-F1 captures the answer similarity well regardless of the underlying workflow step. SummaC, on the other hand, appears inappropriate for matching evaluation (weak negative correlation), but well-suited for alignment. The TRUE score show overall low to moderate alignment with human judgement. In conclusion, the used automatic metrics capture human judgement to some extend and can serve as a solid estimate for performance. The differences in correlation to human judgement for the two different task types in the workflow suggest that a more fine-grained, step-specific evaluation is a promising direction of future research.

\begin{table}
    \centering
    \renewcommand{\arraystretch}{1.3}
    \begin{tabular}{rlll}
         \toprule
         & \textbf{matching} & \textbf{alignment} & \textbf{avg} \\
         \midrule
         \textbf{human} & $0.339$ & $0.5$ & $0.42$ \\
         \textbf{Mixtral} & $0.477$ & $0.667$ & $0.572$\\
         \textbf{GPT-4o} & $0.351$ & $0.75$ & $0.551$\\
         \bottomrule
    \end{tabular}
    \caption{Based on the human evaluation data we compute the Jaccard metric on facts for matching items and compute the average amount of aligned explanations. To aggregate into a single metric, we take the average of both.}
    \label{tab:human_eval_results}
\end{table}

\begin{table}
    \centering
    \renewcommand{\arraystretch}{1.3}
    \begin{tabular}{rll}
         \toprule
         & \textbf{matching} & \textbf{alignment} \\
         \midrule
         \textbf{BERT-F1} & $0.37$ & $0.54$  \\
         \textbf{SummaC} & $-0.22$ & $0.55$ \\
         \textbf{TRUE} & $0.30$ & $0.20$ \\
         \bottomrule
    \end{tabular}
    \caption{Spearman's $\rho$ comparing the human evaluation scores with the automatic metrics for the two scenarios.}
    \label{tab:human_eval_correlation}
\end{table}

\subsubsection{Extended Analysis} \label{ssec:llm_extendedn_experiments}
In this subsection we report on further analysis conducted to gain insights into the reasons for the measured performance of LLMs.

\paragraph{Decision Making} Humans mostly lean towards positive decisions ($67\%$ 'yes' answers), whereas most LLMs are more restrictive ($44\%$-$59\%$ 'yes' answers). A manual answer analysis reveals that human annotators often decide positively despite raising several negative points. They frequently weigh the raised negative points more lightly, reserving a negative decision for extreme cases. LLMs appear to  make more harsh conclusions when the accompanying textual answer contains several points of criticism. This difference in weighing weaknesses in the paper explains the comparatively low performance on decision making for the boolean nodes. Additionally, this points to the risk of a potential negativity bias of LLMs while assessing a paper.

\paragraph{Answer Language} We turn to a language analysis of the textual answers. LLMs consistently generate longer answers ($67$ tokens on average for GPT-3.5) than humans ($55$ average tokens). Upon manual inspection, we find that the automatically generated answers to \textsc{extract} steps tend to split up facts exhaustively, whereas humans make broader and summarizing statements. For \textsc{infer} steps, LLMs often restate the inputs from prior steps, unlike human annotators. Both factors might contribute to the high factual alignment by automatic metrics because LLMs appear more specific and cover many facts, whereas humans tend to select a minimal set of facts relevant to their individual assessment process. This encourages further research on automatic evaluation metrics for this setting that can account for the reasoning context of a step.

\section*{AI-Assistance Disclosure Statement}
We employed LLMs to prepare this paper in a limited and specific capacity. We used LLMs to assist in revising the writing style of this document. Additionally, we employed them as assistants during the coding phase of the experiments. LLMs were \textit{not} involved in the ideation process, the formulation of hypotheses, or the performance of any experiments. All scientific and intellectual contributions are solely the work of the authors.

\end{document}